\documentclass[preprint,12pt]{elsarticle}

\usepackage{amssymb}
\usepackage{multirow}
\usepackage{amsmath, amsthm, amssymb, appendix, bm, graphicx, hyperref, mathrsfs}
\usepackage{algorithm}
\usepackage{algorithmic}
\usepackage{paralist}
\usepackage{makecell}

\usepackage{amsmath}
\usepackage{graphicx}
\usepackage{subcaption}
\usepackage{makecell}
\usepackage{multirow}
\journal{Neurocomputing}

\begin{document}

\begin{frontmatter}

%% Title, authors and addresses

%% use the tnoteref command within \title for footnotes;
%% use the tnotetext command for theassociated footnote;
%% use the fnref command within \author or \affiliation for footnotes;
%% use the fntext command for theassociated footnote;
%% use the corref command within \author for corresponding author footnotes;
%% use the cortext command for theassociated footnote;
%% use the ead command for the email address,
%% and the form \ead[url] for the home page:
% \title{Title\tnoteref{label1}}
% \tnotetext[label1]{}
% \author{Name\corref{cor1}\fnref{label2}}
% \ead{email address}
% \ead[url]{home page}
% \fntext[label2]{}
% \cortext[cor1]{}
% \affiliation{organization={},
%             addressline={},
%             city={},
%             postcode={},
%             state={},
%             country={}}
% \fntext[label3]{}

\title{\textbf{TRACE: \underline{T}ime Se\underline{R}ies P\underline{A}rameter Effi\underline{C}ient Fin\underline{E}-tuning }}

%% use optional labels to link authors explicitly to addresses:
%% \author[label1,label2]{}
%% \affiliation[label1]{organization={},
%%             addressline={},
%%             city={},
%%             postcode={},
%%             state={},
%%             country={}}
%%
%% \affiliation[label2]{organization={},
%%             addressline={},
%%             city={},
%%             postcode={},
%%             state={},
%%             country={}}

\author[a]{Yuze Li\footnote{Corresponding Author. Email: liyuze23@mails.tsinghua.edu.cn}}
\author[b]{Wei Zhu}

% \author{Yuze Li, xx} %% Author name
% %% Author affiliation
\affiliation[a]{organization={Shenzhen International Graduate School, Tsinghua University},%Department and Organization
            city={Shenzhen},
            country={China}}
\affiliation[b]{organization={School of Science, University of Hong Kong},%Department and Organization
            city={Hong Kong},
            country={China}}
% email: wzhu91@connect.hku.hk

%% Abstract
\begin{abstract}
%% Text of abstract
We propose an efficient fine-tuning method for time series foundation models, termed TRACE: Time Series Parameter Efficient Fine-tuning. While pretrained time series foundation models are gaining popularity, they face the following challenges:
(1) Time series data exhibit significant heterogeneity in frequency, channel count, and sequence lengths, necessitating tailored fine-tuning strategies, especially for long-term forecasting.(2) Existing parameter-efficient fine-tuning (PEFT) methods, such as LoRA, are not directly optimized for the unique characteristics of time series data.

To address these challenges, our TRACE framework introduces two key innovations:
(1) Gated DSIC (Gated Dynamic Simulation Importance Calculation), a less biased LoRA module selection mechanism that ensures conditional parameter consistency before and after masking; and (2) Reconstructed forecasting heads for long-term forecasting tasks, which achieve comparable or superior performance to linear probing while drastically reducing parameter counts.

Extensive experiments on long-/short-term forecasting, anomaly detection and natural language tasks across diverse datasets, coupled with ablation studies, validate the effectiveness of our method.
\end{abstract}

% %%Graphical abstract
% \begin{graphicalabstract}
% %\includegraphics{grabs}
% \end{graphicalabstract}

%%Research highlights
% \begin{highlights}
% \item Research highlight 1
% \item Research highlight 2
% \end{highlights}

%% Keywords
\begin{keyword}
%% keywords here, in the form: keyword \sep keyword
fine-tuning\sep time series\sep foundation model\sep forecasting
%% PACS codes here, in the form: \PACS code \sep code

%% MSC codes here, in the form: \MSC code \sep code
%% or \MSC[2008] code \sep code (2000 is the default)

\end{keyword}

\end{frontmatter}

%% Add \usepackage{lineno} before \begin{document} and uncomment 
%% following line to enable line numbers
%% \linenumbers

%% main text
%%

%% Use \section commands to start a section
\section{Introduction}
\label{sec1}
Pre-trained time series foundation models (e.g., MOMENT~\cite{moment}) have demonstrated strong generalization across diverse forecasting and anomaly detection tasks. However, their effective adaptation to downstream applications remains hindered by two key challenges in fine-tuning.

First, standard full fine-tuning is computationally expensive and often infeasible on resource-constrained hardware. In contrast, linear probing—freezing the backbone and tuning only the forecasting head—is efficient but suboptimal: it fails to exploit the rich temporal representations encoded in intermediate layers. This limitation is especially severe in \textbf{long-horizon forecasting}, where the forecasting head itself can contain millions of parameters (e.g., $>35$M for a 720-step horizon in MOMENT), leading to overfitting and degraded performance.

Second, existing parameter-efficient fine-tuning (PEFT) methods like LoRA were designed for natural language tasks and do not account for the unique structure of time series data. Naive application of such methods suffers from \textbf{biased importance estimation} during module selection, resulting in suboptimal adaptation.

To address these issues, we propose \textbf{TRACE} (\textbf{T}ime \textbf{S}eries \textbf{P}arameter \textbf{E}fficient \textbf{F}ine-\textbf{t}uning), a novel framework with two core components:  

\noindent(1) \textbf{Reconstructed forecasting heads} that factorize the original forecasting layer, reducing its parameter count by over 70\% while improving or matching performance. 

\noindent(2) \textbf{Gated DSIC} (\textbf{Gated Dynamic Simulation Importance Calculation}), a debiased LoRA module selection mechanism that simulates post-pruning contexts via Monte Carlo masking to accurately estimate module importance.

Our contributions are summarized as follows:  

\noindent(1) We design lightweight, high-performance forecasting heads tailored for long-horizon time series tasks.  

\noindent(2) We introduce Gated DSIC, a novel PEFT strategy that mitigates evaluation bias and enables dynamic, reversible LoRA pruning.  

\noindent(3) We conduct extensive experiments across long/short-term forecasting, anomaly detection, and even NLP benchmarks, showing that TRACE consistently outperforms full fine-tuning, linear probing, and prior PEFT baselines.  

\noindent(4) We demonstrate that our method generalizes beyond time series, achieving state-of-the-art results on GLUE and QA tasks with minimal trainable parameters.

\section{Related work}
\subsection{Time Series Modeling}

Development of Time Series Forecasting: Time series forecasting has progressed from statistical methods to machine learning, deep learning, Transformer-based architectures, and time series foundation models.

\noindent\textbf{Statistical Methods:} Early methods like Moving Average (MA) and Exponential Smoothing \cite{hyndman2002state} smooth trends but fail to capture complex dynamics. ARIMA \cite{arima} improves accuracy by integrating Autoregressive (AR) and MA components but struggles with nonlinear relationships. VAR \cite{var} models multivariate relationships but suffers from dimensionality issues. These methods are limited to univariate modeling and cannot effectively use covariates.

\noindent\textbf{Machine Learning Methods:} Gradient boosting trees (e.g., XGBoost \cite{chen2016xgboost}, LightGBM \cite{ke2017lightgbm}) leverage feature engineering strategies, such as lag features and sliding window statistics, to model nonlinear patterns. While outperforming statistical methods, their performance depends heavily on feature quality and fixed window lengths limit long-term dependency modeling.

\noindent\textbf{Deep Learning Methods:} RNNs and variants like LSTM \cite{lstm} and GRU \cite{gru} capture temporal dependencies but suffer from inefficiency and vanishing gradients for long-range dependencies. TCNs \cite{tcn} expand receptive fields via dilated convolutions but are constrained by fixed kernels.

\noindent\textbf{Transformer Architectures:} Transformers \cite{vaswani2017attention} excel in capturing global dependencies and enabling parallel computation. Innovations include sparse attention (Informer \cite{zhou2021informer}), channel independence modeling (iTransformer \cite{liu2023itransformer}), frequency-domain enhancement (FEDformer \cite{fedformer}), and causality constraints (PathTST \cite{patchtst}). Challenges remain, including quadratic complexity with sequence length and limited explicit modeling of inherent time series attributes.

\noindent\textbf{Time Series Foundation Models:} Recent foundation models leverage large-scale pretraining on heterogeneous data to learn general representations. Key innovations include multimodal fusion, prompt learning, and support for zero-shot or few-shot forecasting. These models break feature engineering limitations, enhance cross-domain generalization, and handle multivariate/multimodal inputs. Current research focuses on efficient pretraining, interpretability, and lightweight designs.

\subsection{Time Series Foundation Models}
Time Series Foundation Models follow two main approaches: (1) pre-training from scratch on time series data, and (2) adapting large language models (LLMs) for time series tasks.

\noindent\textbf{Approach 1: Pre-training from Scratch.} This approach trains models directly on large-scale time series data (e.g., sensor, financial, or weather data) to capture long-term dependencies, periodicity, and trends using self-supervised tasks like forecasting, imputation, or generation. These models are powerful for time-series tasks but require massive labeled data and high computational costs. Notable examples include TimeGPT \cite{garza2023timegpt}, a Transformer-based model supporting zero-shot transfer; Tempo \cite{cao2023tempo}, a diffusion-based generative model for multivariate time series; TimesFM \cite{timesFM}, a decoder-only architecture with hierarchical time aggregation for irregular data; and Moirai \cite{moirai}, a unified model with multi-task pretraining for long sequences.

\noindent\textbf{Approach 2: Adapting LLMs for Time Series.} This approach maps time series data into textual or other modalities using cross-modal alignment, leveraging LLMs' semantic understanding for tasks like trend description or multimodal reasoning. Challenges include bridging the gap between continuous time series and discrete text, as well as handling causality conflicts with LLM position encodings. Key works include LLMTime \cite{llmTimes}, which discretizes time series into "time tokens" for forecasting via text generation; TEST \cite{sun2023test}, which converts time series into natural language templates for zero-shot forecasting; TimeLLM \cite{jin2023time}, aligning time series with LLMs using embeddings; TEMPO \cite{cao2023tempo}, jointly encoding time series and text for domain-specific tasks; and Lag-Llama \cite{lagLlama}, fine-tuning Llama with lag operators for univariate forecasting.

\section{Preliminaries}
\noindent\textbf{Transformer-based Models.}\;\;  
Transformer \cite{vaswani2017attention} consists of $L$ stacked blocks, each with multi-head attention (MHA) and feed-forward network (FFN) modules. For input $X \in \mathbb{R}^{n \times d}$, MHA computes:
\begin{align}
    &\mathrm{MHA}(X) = \mathrm{Concat}(head_1, \cdots, head_h)W_o,\\
    &\mathrm{head_i} = \mathrm{Softmax}(XW_{q_i}(XW_{k_i})^\top/\sqrt{d_h})XW_{v_i},
\end{align}
where $W_o \in \mathbb{R}^{d \times d}$, $W_{q_i}, W_{k_i}, W_{v_i} \in \mathbb{R}^{d \times d_h}$, and $d_h = d/h$. The FFN uses two linear layers:
\begin{equation}
    \mathrm{FFN}(X) = \mathrm{ReLU}(XW_{f_1}+b_1)W_{f_2}+b_2.
\end{equation}
Residual connections and layer normalization follow both sub-modules. ReLU is often replaced by Gated Linear Units (GLU) \cite{shazeer2020glu}:
\begin{equation}
    \mathrm{GLU}(X) = (XW_{f_0} + b_0) \otimes \sigma(XW_{f_1} + b_1),
\end{equation}
where $\otimes$ is element-wise multiplication and $\sigma$ is the sigmoid function.

\noindent\textbf{Patching.}\;\;  
Time series forecasting benefits from treating subsequences (patches) as tokens rather than individual time points. This approach captures local semantic information and enhances predictive representations \cite{patchtst}.

\noindent\textbf{Channel Independence.}\;\;  
Multivariate time series can be modeled with channel independence, where each token represents data from a single channel \cite{patchtst}. This simplifies the model and allows learning channel-specific dynamics.

\noindent\textbf{MOMENT Model.}\;\;  
MOMENT \cite{moment} is a Transformer-based time series foundation model pretrained on diverse data using masked prediction. It supports long-term/short-term prediction, anomaly detection, and classification. MOMENT-base consists of 12 Transformer encoders with $d = 768$, $h = 12$, and $d_m = 3072$. Its FFN uses GLU:
\begin{equation}
    \mathrm{FFN}(X) = (XW_{f_0} \otimes \mathrm{GELU}(X W_{f_1}))W_{f_2}.
\end{equation}
Input time series ($T = 512$) are divided into $N = 64$ non-overlapping patches ($P = 8$). MOMENT leverages pre-training to capture complex temporal dependencies. We use MOMENT-base to validate TRACE's effectiveness unless otherwise specified.
\section{Our method}

\subsection{ Gated DSIC: Mitigating Bias in LoRA Module }
Parameter-efficient fine-tuning (PEFT) methods like LoRA offer a promising alternative to full fine-tuning. However, naively applying them to time series foundation models is suboptimal. For instance, AdaLoRA parameterizes the weight update via singular value decomposition (SVD) and prunes low-importance singular values to adapt the rank dynamically. While effective in some domains, AdaLoRA faces notable challenges in time-series scenarios. Its SVD-based pruning prioritizes singular values reflecting global feature importance, potentially overlooking critical local signals such as abrupt anomalies or short-term seasonality. Furthermore, its static, irreversible pruning risks underfitting if early-stage pruning discards modules that become relevant later in training.

To address these limitations, we propose Gated Dynamic Simulation Importance Calculation (Gated DSIC), a less biased LoRA module selection mechanism. Gated DSIC introduces a learnable gate for each LoRA module, enabling dynamic and reversible masking. Crucially, it employs a Monte Carlo simulation during the importance evaluation phase to ensure that the importance score of a module is calculated under conditions that mimic the post-masking state, thereby mitigating the bias inherent in sequential, deterministic pruning strategies like AdaLoRA. This ensures conditional parameter consistency before and after masking, allowing the model to recover suppressed modules if needed (e.g., when seasonal patterns re-emerge). Concretely, each LoRA module is associated with a gate whose activation is determined by a less biased importance score estimated via repeated random masking on the validation set. The detailed algorithm is presented in Section \ref{main_algo}. 

\begin{figure}[ht]
    \centering
    \includegraphics[width=1\linewidth]{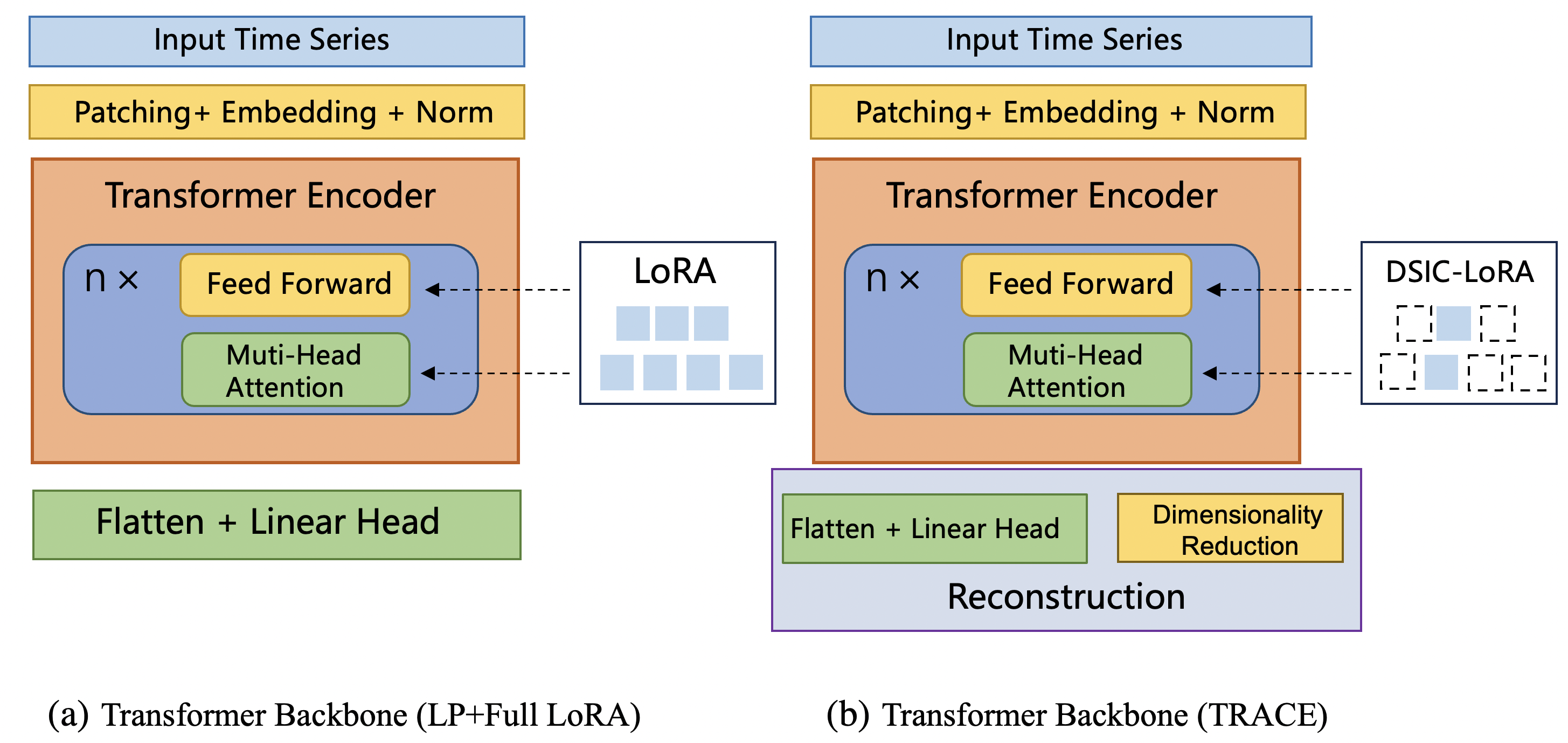}
    \caption{architecture of our method:(a) Represents the default model architecture, which consists of multiple stacked Transformer encoders. LoRA modules are added to all linear layers at each level, and the forecasting head is a linear predictor. Fine-tuning is performed using linear probing and full LoRA adaptation.(b) Represents the TRACE method: LoRA integration follows the Gated DSIC approach, and for long-term forecasting tasks, the forecasting head is dimensionally reduced and reconstructed.}
    \label{architecture}
\end{figure}

Figure \ref{architecture} illustrates the comparison between the TRACE method and the full LoRA fine-tuning model architecture. 
\subsection{Reconstruction of Long-horizon Forecasting}\label{recon}

The standard linear forecasting head in models like MOMENT, parameterized by a matrix $\mathbf{W} \in \mathbb{R}^{(N \times d) \times H}$, becomes prohibitively large for long-horizon forecasting (e.g., $H=720$), often exceeding 35M parameters. While small relative to the full backbone, this introduces significant overfitting risk and computational burden during fine-tuning.

\begin{figure}[htbp]
    \centering
    \includegraphics[width=0.8\linewidth]{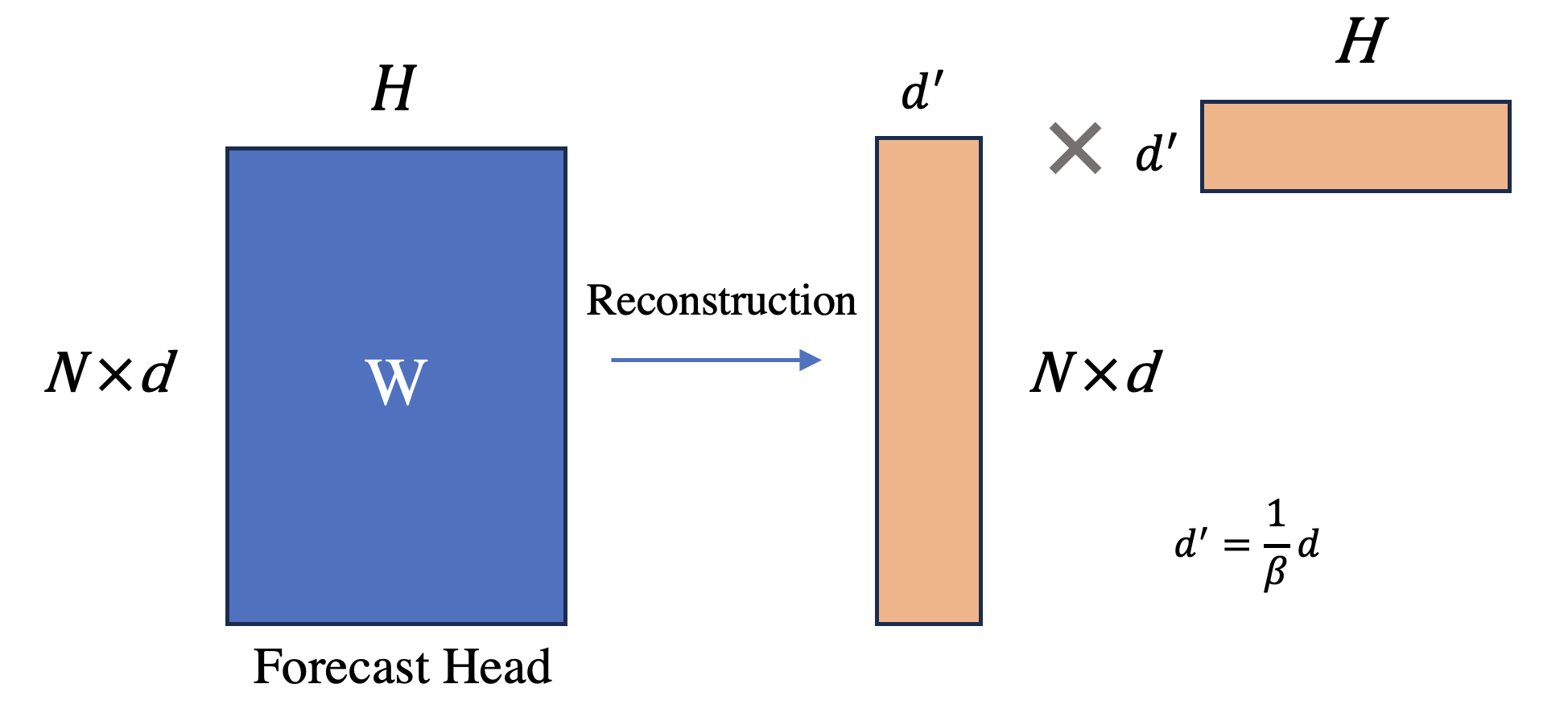} % 假设图片文件名为 forecast_head_recon.png
    \caption{Illustration of forecasting head reconstruction. The original large matrix $\mathbf{W} \in \mathbb{R}^{(N \times d) \times H}$ is factorized into two smaller matrices, reducing the embedding dimension from $d$ to $d' = d/\beta$. This significantly cuts the number of trainable parameters without sacrificing predictive power.}
    \label{fig:forecast_head_recon}
\end{figure}

To address this, we propose reconstructing the forecasting head via low-rank factorization, reducing its parameter count while preserving or enhancing predictive performance. As illustrated in Figure~\ref{fig:forecast_head_recon}, our approach decomposes the original weight matrix $\mathbf{W}$ into two smaller matrices:
\begin{equation}
    \mathbf{W}_{\text{recon}} = \mathbf{W}_1 \cdot \mathbf{W}_2, \quad \text{where} \quad \mathbf{W}_1 \in \mathbb{R}^{(N \times d) \times d'}, \; \mathbf{W}_2 \in \mathbb{R}^{d' \times H},
\end{equation}
and $d' = d / \beta$ is the reduced embedding dimension, controlled by the reduction factor $\beta$.

This reconstruction reduces the parameter count from $\mathcal{O}(N d H)$ to $\mathcal{O}(N d^2 / \beta + d H / \beta)$, achieving reductions of up to 98.2\% (see Table~\ref{tab:beta_reduce}) with minimal performance loss. We experiment with four variants—Projection Down (ProjDown), Feature Truncation (LessFeature), Average Pooling (AvgPool), and 2D Convolution (Conv2D)—each offering different trade-offs between efficiency and representational capacity.

Our experiments (Section 5.5.2) show that ProjDown and AvgPool consistently outperform the baseline linear probing, validating the effectiveness of our reconstruction strategy.

\begin{table}
\centering 
\caption{
% 不同预测horizon下，模型预测头的参数量以及 $\beta$不同取值的条件下预测头的参数减小量及减小百分比
Comparison of forecasting head parameters across different horizons and reduction under varying $\beta$ values
}
\begin{tabular}{ccccc}
\hline
\multicolumn{1}{l}{Horizon} &$\beta$&  Head Parameters & Reduced & Reduction Percentage   \\
\hline
\multirow{3}{*}{192}  & 4 & 9,437,184& 2,506,752 & 73.4\% \\
                      & 8 & 9,437,184 & 1,253,376 & 86.7\% \\
                      & 16 & 9,437,184 & 626,688 & 93.4\% \\
                      \hline
\multirow{3}{*}{720}  & 4 & 35,389,440 & 8,994,816 & 74.6\% \\
                      & 8 & 35,389,440 & 4,497,408 & 87.3\% \\
                      & 16 & 35,389,440 & 626,688 & 98.2\% \\
\hline
\end{tabular}
\label{tab:beta_reduce}
\end{table}

\ref{head} is a concise description of the reconstruction operations applied to the forecasting head. The parameter $d$ represents the embedding dimension for each patch (subsequence), with $d=768$ in MOMENT. To leverage the pre-trained model's predictive capabilities and reduce overfitting, we reduce $d$ since patch length limits information content.

\subsection{Our Gated DSIC method}\label{main_algo}
\subsubsection{Learnable Gating for Dynamic LoRA Control}
\label{chap:gate}
We propose a method to identify the most important and effective LoRA modules, allowing for efficient fine-tuning within a limited parameter budget.
To enable dynamic and reversible selection of LoRA modules during parameter-efficient fine-tuning, we introduce \textbf{a learnable scalar gate} \( g_{ij} \in \mathbb{R} \) for the \(j\)-th LoRA module in the \(i\)-th Transformer layer. This gate multiplicatively modulates the low-rank update, yielding the adapted layer output:
\[
h_{ij} = W^{(0)}_{ij} x + g_{ij} \cdot \Delta x_{ij} = W^{(0)}_{ij} x + g_{ij} \cdot BAx,
\]
where \( W^{(0)}_{ij} \) denotes the pre-trained weight matrix, and \( A \in \mathbb{R}^{r \times d_2} \), \( B \in \mathbb{R}^{d_1 \times r} \) are the low-rank adaptation matrices with rank \( r \ll \min(d_1, d_2) \). The gate \( g_{ij} \) is initialized to 1 and optimized jointly with \( A \) and \( B \) via backpropagation.

Crucially, the importance of each LoRA module is estimated through the magnitude of the gradient of the validation loss with respect to its gate:
\begin{equation}
    I(g_{ij}) = ||\nabla_{g_{ij}} \mathcal{L}_{val}||
\end{equation}
This gradient reflects how sensitive the model’s performance is to perturbations in \( g_{ij} \), providing a natural proxy for module contribution. 

Taking MOMENT-base as an example, which consists of 12 Transformer encoder layers with 7 linear sub-layers each (Query, Key, Value, Output, and three FFN projections), this yields a total of 84 gates, denoted as \( \mathcal{G} = \{ g_{ij} \mid i=1,\dots,12;\, j=1,\dots,7 \} \).

Once we have $I(g_{ij})$, we can prune the parameters of those LoRA modules that contribute less to the fine-tuning task. This helps reduce the parameter count and prevent overfitting.

Following common practice, we define the importance score of unmasked gates as the expected sum of gradient magnitudes:
\begin{equation}
    I(g_{ij}) = \mathbb{E}_{(X,Y) \sim \mathcal{D}_{\text{val}}} \left[ \sum_{t=1}^T \left| \frac{\partial \mathcal{L}_{\text{val}}}{\partial g_{ij}} \right| \right]
\end{equation}
\textbf{However, this method has a bias in evaluating importance.} For example, when calculating the importance $I(g_{kl})$, the implicit assumption is that the effect of a small fluctuation $\Delta g_{kl}$ on the overall loss is measured while keeping $|g_{ij}| = 1$ for all other $i \neq k\;and\;j \neq l$. However, when we mask some of the LoRA Gates based on their importance rankings, the premise of this assumption changes, leading to bias in the importance evaluation. Specifically, for $g_{kl}$, assuming it remained unmasked and some other LoRA Gates were masked in one operation, we observe:
\begin{equation}
    I(g_{kl}| \;g_{ij}=1,g_{ij}\in \mathcal{G})\ne I(g_{kl}| \;g_{mn}=0,g_{mn}\in 
\tilde{\mathcal{G}}\;and\;g_{pq}=1,g_{pq}\notin\tilde{\mathcal{G}})
    \label{proof}
\end{equation}
where $\tilde{\mathcal{G}}$ denotes the set of masked LoRA Gates. Equation (\ref{proof}) demonstrates that the masking of some LoRA Gates will change the importance of the remaining unmasked LoRA Gates, thus introducing bias.

\subsubsection{Our solution: Gated Dynamic Simulation Importance Calculation}
As discussed in Section \ref{chap:gate}, the standard gradient-based importance evaluation (e.g., as used in AdaLoRA) is biased. This bias arises because the importance score for a LoRA module is calculated under the assumption that all other modules are active. However, during pruning, many other modules are masked (set to zero), which fundamentally alters the model's context and thus the true contribution of the remaining modules. This inconsistency between the evaluation condition (all modules active) and the deployment condition (many modules masked) leads to an inaccurate and biased importance estimate.

To address this, we propose the Gated Dynamic Simulation Importance Calculation (Gated DSIC) method, which aims to reduce this evaluation bias by simulating the post-pruning context during the importance calculation phase. The core idea is to evaluate a module's importance not in isolation, but within a dynamically changing coalition of other active and masked modules, thereby better approximating its true contribution in the final pruned model.

Specifically, our method performs a Monte Carlo simulation on the validation set. In each simulation trial, we randomly mask a subset of the LoRA gates. The importance of the remaining unmasked gates is then computed based on their gradients in this specific masked context. By repeating this process many times (M times) and averaging the importance scores across all trials, we obtain a more robust and less biased estimate of each module's contribution. This procedure ensures that the importance evaluation condition is consistent with the diverse contexts the module will experience after pruning, leading to a more reliable selection of modules to retain.

The superiority of our approach in reducing bias is empirically validated by comparing its importance scores against the gold-standard Shapley Value method, as proposed in \cite{held2022shapley}. Following the protocol in \cite{held2022shapley}, we compute the Shapley Values for each LoRA module in a MOMENT-base model fine-tuned on the ETTh1 dataset. We then compare the correlation between these Shapley Values and the importance scores generated by our Gated DSIC method and the gradient-based method used in AdaLoRA. As shown in Table \ref{tab:correlation_shapley} , our method achieves a 63.3\% correlation with the Shapley Values, while AdaLoRA's method shows a significantly lower correlation of -22.1\%. This strong alignment with the theoretically sound Shapley Value method demonstrates that Gated DSIC provides a much less biased estimation of module importance compared to conventional gradient-based approaches. This allows our method to achieve performance close to the Shapley-based selection while being computationally feasible (0.25 GPU hours vs. 27 GPU hours for Shapley computation).

\begin{figure}[htbp]
    \centering
    \includegraphics[width=0.8\linewidth]{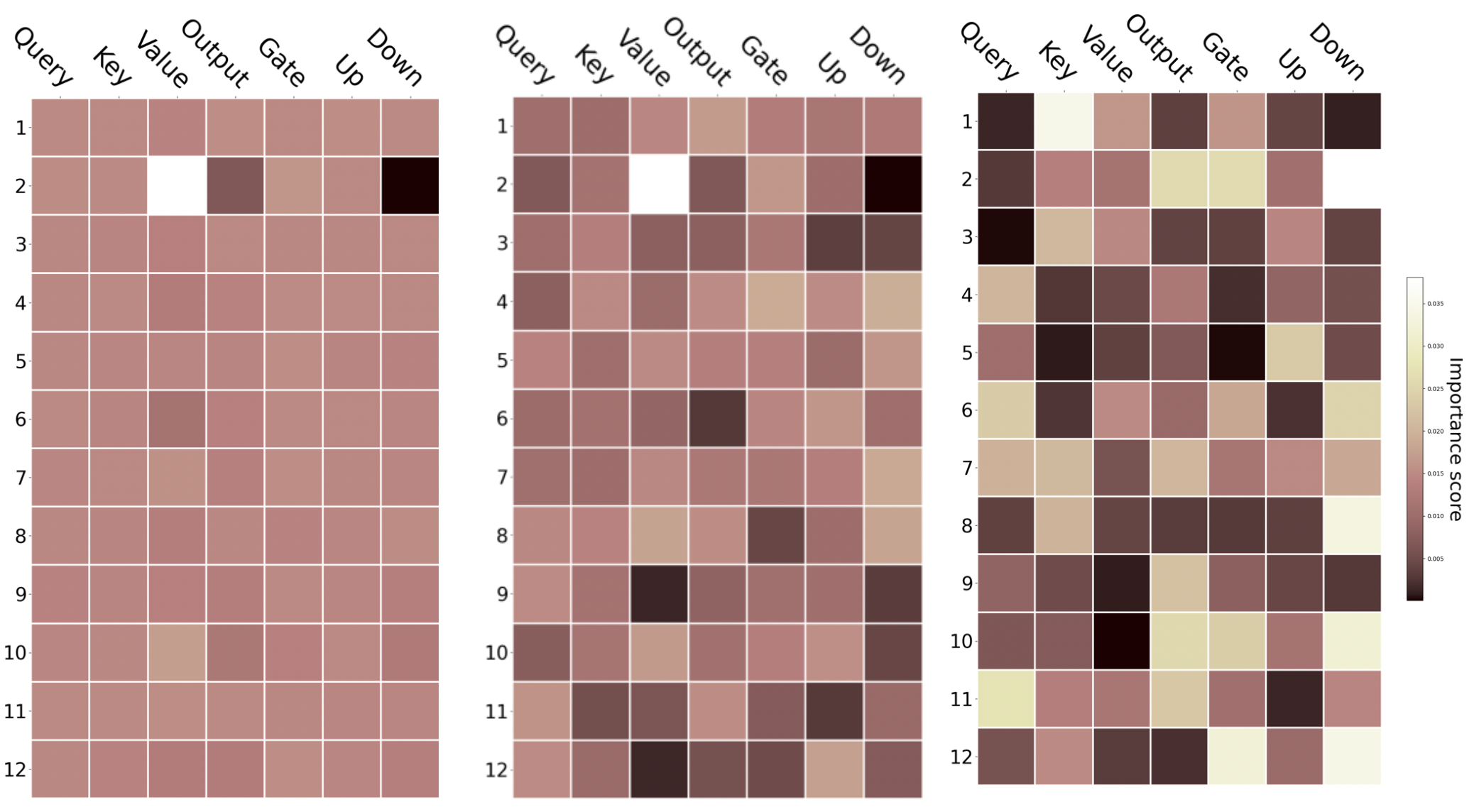}
    \caption{Visualization of LoRA Module Importance Scores.\\
Left: AdaLoRA (Gradient-based). Middle: Gated DSIC (Ours). Right: Shapley Value. The color intensity represents the magnitude of the importance score, with darker shades indicating higher importance. The x-axis represents different LoRA module types (Query, Key, Value, Output, Gate, Up, Down), and the y-axis represents the Transformer layer number (1-12).
}
    \label{fig:shapely}
\end{figure}

\begin{table}[htbp]
\centering
\caption{Correlation of Importance Scores with Shapley Values. We compute the Spearman correlation between the importance scores from different methods and the ground-truth Shapley Values for LoRA modules in a MOMENT-base model fine-tuned on ETTh1.}
\label{tab:correlation_shapley}
\begin{tabular}{lc}
\hline
METHOD& CORRELATION WITH SHAPLEY VALUE \\
\hline
Gated DSIC (Ours) & 63.3\% \\
AdaLoRA (Gradient-based) & -22.1\% \\
\hline
\end{tabular}
\end{table}
To further illustrate this point, we visualize the LoRA module importance scores calculated by the three methods on a MOMENT-base model fine-tuned on ETTh1. As shown in Figure \ref{fig:shapely}, the importance distribution generated by AdaLoRA (left) is highly concentrated, indicating a potential bias towards a small subset of modules. In contrast, the distribution from our Gated DSIC method (middle) exhibits a more diverse and balanced pattern, closely resembling the distribution produced by the Shapley Value method (right), which is considered the gold standard for credit assignment. This visual evidence corroborates our quantitative findings, confirming that Gated DSIC effectively mitigates the evaluation bias inherent in standard gradient-based pruning. 

These quantitative and qualitative analyses confirm that Gated DSIC effectively mitigates the evaluation bias inherent in standard gradient-based methods. The detailed procedure of our method is as follows:

\noindent\textbf{Stage 1: Joint Training}

We simultaneously optimize the low-rank matrices $\{A, B\}$, reconstruction forecasting head $H$ and the gating parameters $\mathcal{G} = \{g_{ij}\}$ on the training set. The loss function is:
\begin{equation}
    \mathcal{L}_{\text{train}} = \mathcal{L}_{\text{task}}(f(X; W_0 + g \cdot BA,H), Y)
\end{equation}

\noindent\textbf{Stage 2: Less Biased Importance Evaluation}

We perform a Monte Carlo masking experiment on the validation set:
\begin{enumerate}
    \item \textbf{Random Masking}: With probability \( p \), randomly set some of the gates to 0, i.e., \( g_{ij} \leftarrow 0 \), and denote the masked set as $\tilde{\mathcal{G}} \subseteq \mathcal{G}$. The remaining gates are kept as 1. This operation is counted as $k$.
    \item \textbf{Importance Calculation}: We compute the importance of the gates on the validation set $\mathcal{D}_{val}$:
    \begin{equation}
    I^{(k)}(g_{ij}) = 
    \begin{cases} 
    \mathbb{E}_{(X,Y) \sim \mathcal{D}_{\text{val}}} \left[ \sum_{t=1}^T \left| \frac{\partial \mathcal{L}_{\text{val}}}{\partial g_{ij}} \right| \right] & \text{if } g_{ij} \notin \tilde{\mathcal{G}} \\
    0 & \text{if } g_{ij} \in \tilde{\mathcal{G}}
    \end{cases}
    \label{imp}
     \end{equation}
    \item \textbf{Repeated Sampling}: Repeat the random masking operation \( M \) times, each time obtaining a different masked set $\tilde{\mathcal{G}}$ and corresponding importance scores $I^{(k)}(g_{ij})$.
    \item \textbf{Average Gradient}: Take the average gradient importance across the \( M \) samples to obtain a less biased and stable evaluation:
    \begin{equation}
        \bar{I}(g_{ij}) = \frac{1}{M} \sum_{k=1}^M  I^{(k)}(g_{ij}) \label{eq:cal_imp}
    \end{equation}
\end{enumerate}

\begin{figure}[htbp]
    \centering
    \includegraphics[width=1\linewidth]{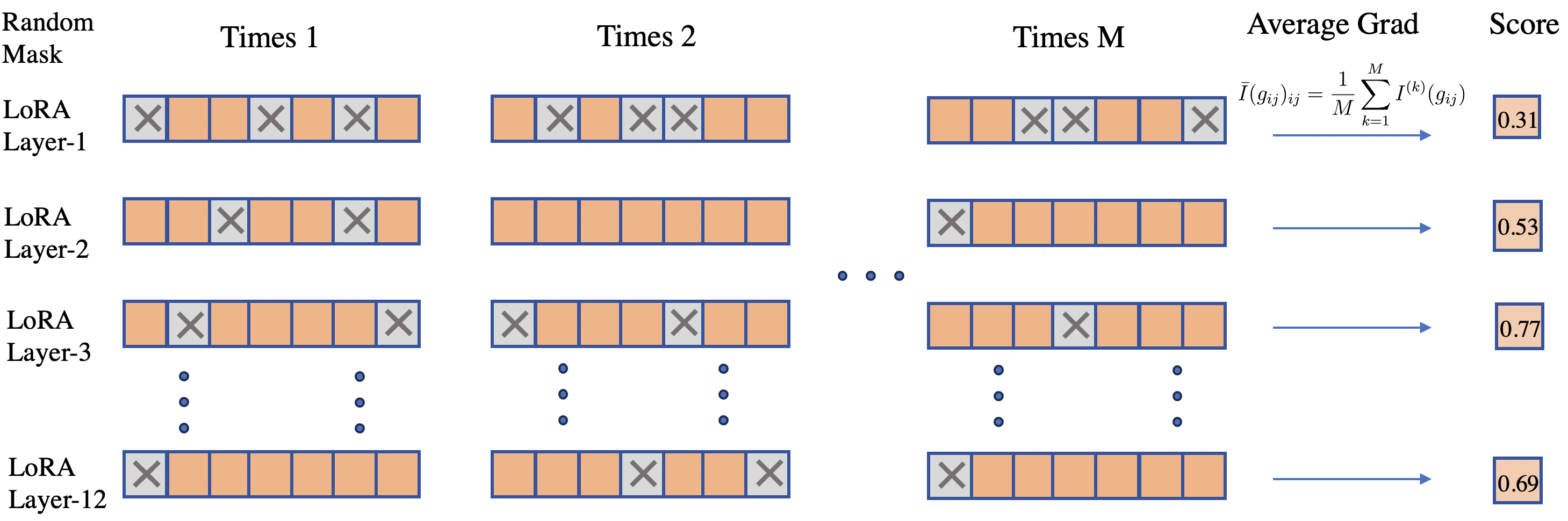}
    \caption{Illustration of the Gated DSIC Importance Scoring Process}
    \label{fig:mask1}
\end{figure}

 Figure \ref{fig:mask1} visualizes how our method calculates the importance score for each LoRA gate. In each of $M$ independent trials, a random subset of LoRA gates is masked (indicated by grey boxes with '×'). The gradient magnitude of the unmasked gates is then computed on the validation set. The final importance score for each gate (e.g., 0.31, 0.53) is obtained by averaging its gradient magnitudes across all M trials, as defined in Equation \ref{eq:cal_imp}. This Monte Carlo simulation ensures that the evaluation context approximates the post-pruning state.

\noindent\textbf{Gate Pruning}

Based on $\bar{I}(g_{ij})$, we rank the gates and prune the $p\%$ gates with the lowest importance (set them to 0), thereby retaining the most effective LoRA modules. This process is repeated iteratively over training epochs.

\begin{figure}[htbp]
    \centering
    \includegraphics[width=1\linewidth]{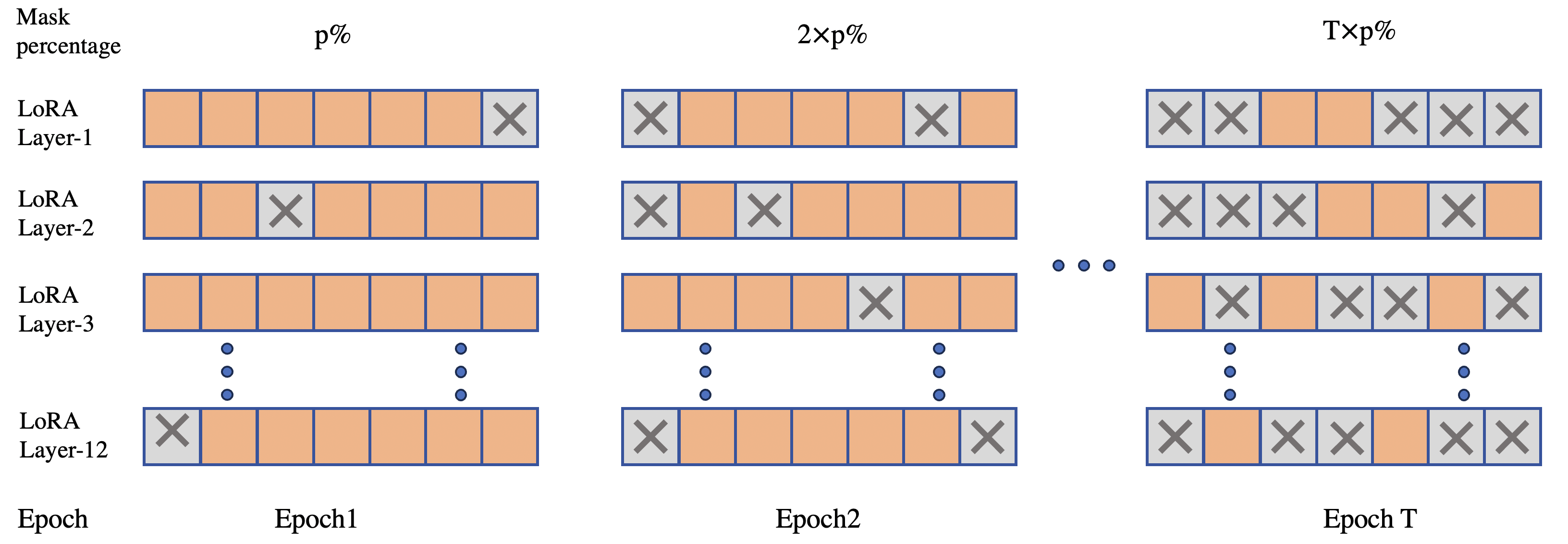}
    \caption{Illustration of the Iterative Pruning Schedule}
    \label{fig:mask2}
\end{figure}

Figure \ref{fig:mask2} depicts the iterative pruning process over $T$ epochs. At each epoch $t$, we prune an additional $p\%$ of the currently unmasked LoRA gates based on their importance scores from Figure \ref{fig:mask1}. This continues until the total number of pruned gates reaches the predefined budget (e.g., $T\times p\%$), achieving efficient fine-tuning with minimal parameter usage.

This method, through multiple random experiments, constructs model contexts that approximate the post-pruning state, leading to a more reliable and less biased importance estimation for LoRA gates. 

\subsubsection{Workflow}

We present the complete calculation workflow for our Gated DSIC method:

\begin{algorithm}[H]
    \caption{The whole workflow of our Gated DSIC method}
    \begin{algorithmic}[1]
    \REQUIRE Train dataset $\mathcal{D}_{train}$, Validation dataset $\mathcal{D}_{val}$; total iterations $T$; budget schedule $\{b^{(t)}\}^T_{t=0}$; hyperparameters $p$
    \FOR{t=1,\dots,T}
    \STATE First, train for $\alpha$ steps on the training dataset $\mathcal{D}_{train}$
    \FOR{k=1,\dots,M}
    \STATE Random Masking: Randomly set $p\%$ of the currently unmasked LoRA gates to 0, denoted as $\tilde{\mathcal{G}}$
    \STATE Sample a batch from $\mathcal{D}_{val}$ and compute the gradient $\nabla \mathcal{L}(g,\mathcal{A,B})$
    \STATE Compute the sensitivity $I^{(k)}(g_{ij})$ for the LoRA gates $g_{ij} \in \mathcal{G}/\tilde{\mathcal{G}}$
    \ENDFOR
    \STATE Compute the average sensitivity for all LoRA gates: $\bar{I}(g_{ij}) = \frac{1}{M} \sum_{k=1}^M I^{(k)}(g_{ij})$ and mask the top $p\%$ ranked gates
    \STATE Continue training for $\alpha$ steps, then repeat the process of masking the top $p\%$ of LoRA gates until the parameter budget $b$ is reached
    \ENDFOR
    \ENSURE $\tilde{\mathcal{G}}$
    \end{algorithmic}
    \label{workflow}
\end{algorithm}

Algorithm \ref{workflow} describes the Gated DSIC method. The procedure begins by dividing the data into a training set $\mathcal{D}_{train}$ and a validation set $\mathcal{D}_{val}$. After training for $\alpha$ steps on the training set, we proceed with the following steps on the validation set: first, we randomly mask $p\%$ of the LoRA gates, then compute the sensitivity for the unmasked LoRA gates using equation(\ref{imp}). After repeating the experiment $M$ times, we calculate the average sensitivity for each LoRA gate and mask the top $p\%$ gates. This process is repeated until the number of masked LoRA modules reaches the predefined parameter budget $b$. Finally, the method outputs the set of LoRA gates to be masked, $\tilde{\mathcal{G}}$.

We have evaluated the Gated DSIC method on these tasks: long-term time series forecasting, short-term forecasting, anomaly detection and natural language processing.

\section{EXPERIMENTS}
\subsection{Datasets and Evaluation Metrics}
\noindent\textbf{Datasets.} For forecasting and anomaly detection, we used subsets of the MOMENT dataset. Long-term forecasting tasks include ETTh1, ETTh2, ETTm1, ETTm2 \cite{etth}, Weather, and Exchange \cite{zhou2021informer}. Short-term forecasting uses M4 and M3 from the Monash archive \cite{godahewa2021monash}, consisting of Yearly (macroeconomic indicators, 5–50 years, horizon=6), Quarterly (sales/economic data, 10–100 quarters, horizon=8), and Monthly (retail/energy/weather data, 50–1000 months, horizon=18). Anomaly detection uses the TSB-UAD benchmark \cite{tsb_uda}, with 1980 univariate time series annotated from 18 public datasets. We selected 9 series from MITDB, ECG, MGAB, and SVDB for experiments.

\noindent\textbf{Metrics.} We used multiple metrics commonly employed in task-specific benchmarks to evaluate each experiment, including MSE and MAE for long-term forecasting, and sMAPE for short-term forecasting. The vanilla F1 score is not suitable for time series tasks, as it ignores their sequential nature. Therefore, for anomaly detection performance, we use the widely accepted adjusted best F1 score.

\begin{table}
\centering 
\caption{Dataset Splits, Number of Channels, and Prediction Lengths for Different Tasks and Datasets}
\scalebox{0.8}{ % 0.9 表示缩放比例
    {\normalsize
\begin{tabular}{c|c|c|c|c}
\hline
\multicolumn{1}{l|}{Task} & Dataset & Channels & Series Length & Data Size (Train, Val, Test)  \\
\hline
\multirow{4}{*}{\parbox{2cm}{Long horizon forecasting\\ (Informer)}}  & ETTh1,ETTh2 & 7& \multirow{3}{*}{\{192,720\}} & (8033, 2785, 2785) \\
                      & ETTm1,ETTm2 & 7 &  & (33953, 11425, 11425) \\
                      & Weather & 21 &  & (36280, 5175, 10444) \\
                      &Exchange&8&\{96,192\}&(4704, 665, 1422)\\
                      \hline
\multirow{6}{*}{\parbox{2cm}{Short horizon forecasting (Monash)} } & M4-Yearly & 1 & 6 & (16099, 2301, 4600) \\
                      & M4-Quarterly & 1 & 8 & (16800, 2400, 4800) \\
                      & M4-Monthly & 1 & 18 & (33600, 4800, 9600) \\
                      & M3-Yearly&1&6
                      &(451, 65, 129)\\
                      & M3-Quarterly&1&8
                      &(529, 76, 151)\\
                      &M3-Monthly&1&18
                      &(999, 144, 285)\\

\hline
\multirow{4}{*}{\parbox{2.3cm}{Anomaly \quad detection (TSB-UAD)} } & MITDB &1  &  \multirow{4}{*}{-}&  \multirow{4}{*}{(50\%, -, 50\%)}\\
                      & ECG & 1 &  &  \\
                      & MGAB & 1 &  &  \\
                      &SVDB&1&&\\
\hline

\end{tabular}}}
\label{split}
\end{table}
\subsection{Compared models}
All of our models are based on the MOMENT-base pre-trained model. For the forecasting task, we compare the performance of MOMENT under four different modes: linear probing, full LoRA fine-tuning, reconstructed forecasting head, and TRACE fine-tuning. The details of these methods are as follows:

\noindent \textbf{Forecasting: } 
\textbf{(1)} \textit{Linear Probing:} The model is linear-probed (\( \text{MOMENT}_{LP} \)) by freezing all parameters except for those in the reconstruction or forecasting head.  
\textbf{(2)} \textit{LoRA Fine-tuning:} Based on linear probing, LoRA modules are added to all linear layers for fine-tuning.  
\textbf{(3)} \textit{Reconstructed Forecasting Head:} The forecasting head is reconstructed using the four methods in sec \ref{recon}: $Proj\;Down, Avg\;Pool, Less\;Feature$ and $Conv\;2D.$ \textbf{(4)} $AdaLoRA$: $Proj Down$ forecasting head with AdaLoRA method.
\textbf{(5)} \textit{TRACE:} In addition to the reconstructed forecasting head, the Gated DSIC method is incorporated. The main model for TRACE uses the $Proj\;Down$ method for dimensionality reduction in the forecasting head.

\noindent \textbf{Anomaly Detection: }
\textbf{(1)} \textit{Linear Probing:} The model is linear-probed (\( \text{MOMENT}_{LP} \)) by freezing all parameters except for those in the reconstruction or forecasting head.  
\textbf{(2)} \textit{LoRA Fine-tuning:} Based on linear probing, LoRA modules are added to all linear layers for fine-tuning.  
\textbf{(3)} \textit{TRACE:} The DSIC-LoRA method is applied on top of Linear Probing for anomaly detection.

\subsection{Settings}
% \subsubsection{设备}
\subsubsection{model hyper-parameters} 
The hyperparameters used for training our models are shown in the Table \ref{para}:
\begin{table}[H]
\caption{
Main experiment parameter settings of the TRACE method
}
\centering \scalebox{0.9}{ % 0.9 表示缩放比例
    {\normalsize
\begin{tabular}{c|l|c}
\hline
Model & Hyper-parameters&Value   \\
\hline
\multirow{11}{*}{MOMENT(TRACE)} 
                    &dimension of model&768\\
                    &number of layers&12\\
                    &number of heads&12\\
                    & Sequence length &512\\
                      & patch length &8\\
                      &dimension of feedforward layer&3072\\
                      &lora rank $\gamma$ &2\\
                      & max rank of AdaLoRA & 4\\
                      &total mask percentage& 95\%\\
                      &mask percentage by step&10\%\\
                      &random masking operation times $M$&8\\
                      &forecast horizon&\{192,336,720\}\\
                      &Dimensionality Reduction Factor $\beta$&8\\
                      
                      \hline
\end{tabular}}}

\label{para}
\end{table}

\subsubsection{training hyper-parameters} 

For the long-term forecasting task, during the model training phase, we use a backtracking window of length \( L = 512 \) for training all models. We predict the Exchange dataset with \( T = 96 \) and \( T = 192 \) time steps, while for the other datasets, the prediction horizons are set to \( T = 192 \) and \( T = 720 \) time steps. We evaluate the models using Mean Squared Error (MSE) and Mean Absolute Error (MAE) as metrics. 

The initial learning rate during training is set to \( 2 \times 10^{-4} \), with a maximum of 15 epochs for training. The validation and test sets are split according to the rules outlined in Table \ref{split}, and we adopt the OneCycleLR strategy for early stopping. The batch size used during model training is 16.

For the short-term forecasting task, we follow the settings defined in MOMENT. For the Monthly dataset, the prediction horizon \( T = 18 \); for the Quarterly dataset, \( T = 8 \); and for the Yearly dataset, \( T = 6 \).

For the anomaly detection task, the training parameters remain consistent with those used in the forecasting tasks.

\subsection{Main Experimental Results}
\subsubsection{Long-horizon Forecasting}
\begin{table}[H]
\centering
\caption{Long-term forecasting performance measured using Mean Squared Error (MSE) and Mean Absolute Error (MAE).Results are averaged over five random experiments. Bold and underlined values indicate the best and second-best results, respectively.}
\scalebox{0.67}{ % 0.9 表示缩放比例
    {\normalsize
\begin{tabular}{cc|cc|cc|cc|cc|cc}
\hline
\multicolumn{2}{c}{\multirow{2}{*}{ \makecell{Forecasting \\Horizon}}} & \multicolumn{2}{c}{$\;\;\mathrm{MOMENT}_{LP}\;\;$}    & \multicolumn{2}{c}{$\mathrm{MOMENT}_{LP+LoRA}$} & \multicolumn{2}{c}{$\mathrm{MOMENT}_{Proj\;down}$}&\multicolumn{2}{c}{$\mathrm{MOMENT}_{AdaLoRA}$} & \multicolumn{2}{c}{\;\;\;\;\;TRACE\;\;\;\;\;} \\

\multicolumn{2}{c}{}& \multicolumn{1}{|c}{MSE} & MAE & \multicolumn{1}{|c}{MSE}  & MAE  & \multicolumn{1}{|c}{MSE}     & MAE   & \multicolumn{1}{|c}{MSE}       & MAE   & \multicolumn{1}{|c}{MSE}       & MAE        \\ \hline

\multirow{3}{*}{ETTh1}              & 192           &  0.417  & 0.430    &      0.420                    & 0.433     &  0.415                           & 0.430       & \underline{ 0.414}  &\underline{0.429} & \textbf{0.412}                           & \textbf{0.427}                                      \\
                                    & 336           & 0.432                        &  0.449   &   0.435                       &  0.450    &   \underline{0.429}                          &\underline{0.446}         &0.430 &\underline{0.446} & \textbf{0.428}                              &   \textbf{0.444}                                          \\
                                    & 720           &0.476    &0.487     &0.479                          &0.490      & 0.475                            &   0.486     & \underline{0.472}&\underline{0.484}  &  \textbf{0.470}                             & \textbf{0.481}            \\ \hline
\multirow{3}{*}{ETTh2}              & 192           & 0.347                        &  0.389   &0.348                          &  0.388    & 0.347                            & 0.386        &    \textbf{0.345}    & \textbf{0.384}&           \underline{0.346}            &    \underline{0.385}                                       \\
                                    & 336   &   0.374                      & 0.413    & 0.376                         &0.417      &\underline{ 0.373 }                           &\underline{0.412}       & 0.376& 0.416 &      \textbf{0.370}                         & \textbf{0.409}                                      \\
                                    & 720           &0.397                         & 0.436    &  0.398                        &  0.437    &  \underline{0.396}                           &\underline{0.435}        &0.399&0.437 &  \textbf{0.395}                             &      \textbf{0.434}                                    \\ \hline
\multirow{3}{*}{ETTm1}              & 192           &   0.328                      &  0.368   &0.329                          & 0.371     & \underline{0.327}                            & \underline{0.367}     & 0.330&0.372  &       \textbf{0.326}                        & \textbf{0.366}                                       \\
                                    & 336           &0.355                         & 0.388    &   0.356                       & 0.389     &0.353                             & 0.387        &   \underline{0.352}&\underline{0.386}&        \textbf{0.351}                    &        \textbf{0.384}                                  \\
                                    & 720           & 0.408                        &  0.419   &   0.411                       & 0.422     &         \underline{0.406}                    & \underline{0.417}        &  0.407&0.418&\textbf{0.405}                             &           \textbf{0.416}                               \\ \hline
\multirow{3}{*}{ETTm2}              & 192           &    0.228                     &  0.299   &                    0.229     &  0.300    & \textbf{0.226}                            &   \textbf{0.297}      &        \underline{0.227}     &\underline{0.298}&        0.228          &    0.299                                    \\
                                    & 336           & 0.276                        &  0.329   &  0.279                        & 0.331     &   0.276                          &0.328         &     \underline{0.275}&\underline{0.327}& \textbf{0.274}                          &                        \textbf{0.325}                \\
                                    & 720           &  0.360                       & 0.386    &  0.362                        & 0.387     &    \underline{0.360}                          &  \underline{0.384}       &0.363 &0.388& \textbf{0.359}                             & \textbf{0.383}                                      \\ \hline
\multirow{3}{*}{Weather}            & 192           &     0.204                    &0.254     &                   0.207       & 0.258     &                    \underline{0.203}         & \underline{0.252}        &                    0.206 &0.257&    \textbf{0.200}      &     \textbf{0.248}                                     \\
                                    & 336           &0.253                         & 0.294    &0.255                          & 0.296     &          \underline{0.252}                   &   \underline{0.293}      & 0.254&0.296&\textbf{0.251}                              &  \textbf{0.292}                                     \\
                                    & 720           &   0.323                      &  0.340   &  0.324                        & 0.340     & 0.321                            & 0.339        &  \underline{0.318} &\underline{0.338}&\textbf{0.317}                             &    \textbf{0.337}                                     \\ \hline
\multirow{3}{*}{Exchange}           & 96            &    0.122                     &  0.249   &               0.119           &  0.247    & \textbf{0.105}                            &   \textbf{0.230}    &0.111&0.237  &                            \underline{0.108}   &    \underline{ 0.233}                                     \\
                                    & 192           & 0.225                        &  0.342   & 0.222                         &0.340      &  0.214                          & 0.333        &    \underline{0.210}&\underline{0.329}&\textbf{0.206}                           &    \textbf{0.326}                                   \\
                                    & 336           &     0.317                    &0.434     &  0.320                        &  0.437    &           \underline{0.315}                  &    \underline{0.433}     &           0.316&0.434&\textbf{0.313}                    &             \textbf{0.430}                            \\
                                    \hline
            \multirow{3}{*}{ECL}
                                     & 192            &    0.158                     &  0.254   &               0.160           &  0.255    & 0.153                            &   0.248      &                   \textbf{0.149}  &\textbf{0.244}&       \underline{0.151}   &     \underline{0.246}                                     \\
                                    & 336           & \underline{0.169}                       &  \underline{0.266}   & 0.172                         &0.268      &  0.170                          & 0.267        &    0.171&0.267&\textbf{0.167}                           &    \textbf{0.264}                                   \\
                                    & 720           &    0.210                   &0.301     &  \underline{0.209}                        &  \underline{0.300}    &          \textbf{0.208}                  &    \textbf{0.298 }  &0.210& 0.301   &           \textbf{0.208}                    &             \textbf{0.298 }                            \\
                                    \hline
                                    
\end{tabular}}}
\label{main_exp}
\end{table}

Table \ref{main_exp} presents the main experimental results for long-term forecasting, comparing the performance of the MOMENT model under five conditions: linear probing, full LoRA fine-tuning, reconstruction of the forecasting head (projection down, with $\beta=8$), AdaLoRA(with projection down head) and TRACE(Gated DSIC and projection down with $\beta=8$) fine-tuning. The results are evaluated in terms of MSE and MAE across different prediction lengths. 

By comparing the results, we observe that the reconstructed forecasting head: $\mathrm{MOMENT}_{Proj\;down}$, achieves lower MSE and MAE across all long-term forecasting tasks compared to linear probing ($\mathrm{MOMENT}_{LP}$) and full LoRA fine-tuning($\mathrm{MOMENT}_{LP+LoRA}$). Our proposed \textbf{TRACE}  outperforms other approaches, including AdaLoRA, on the majority of datasets. 

\textbf{Static Test: }
We further conducted a statistical significance test (McNemar’s test, p-value=0.0071) on the prediction errors, which confirms that TRACE’s improvements over $\mathrm{MOMENT}_{LP}$ are statistically significant.

The detailed methodology and results of this test are provided in Appendix \ref{apx:Statistical}

\textbf{Computational Cost: } To address concerns regarding the computational overhead of our Gated DSIC importance evaluation, we provide a detailed breakdown of training and inference costs on the Electricity dataset (prediction horizon = 720, $\beta$ = 4) in Table \ref{macs}. Our TRACE method requires 942.86 seconds of training time, which is approximately 1.30× that of full LoRA (729.58s) and 1.47× that of AdaLoRA (643.31s). This modest increase in training time is solely attributed to the offline Monte Carlo importance evaluation on the validation set and is deemed acceptable given the consistent performance gains demonstrated in Table \ref{main_exp}. Crucially, the inference time of the final pruned TRACE model (38.17s) is nearly identical to that of AdaLoRA (37.95s) and full LoRA (38.36s), confirming that our method imposes no additional latency during deployment. Furthermore, TRACE achieves the lowest backbone parameter count (2.89M) among all PEFT baselines, enhancing its efficiency for resource-constrained scenarios. In summary, the slight offline training overhead is a worthwhile trade-off for a more effective and deployment-ready fine-tuned model. 
% Table \ref{macs}  presents the computational costs of different models on the Electricity dataset, along with experimental results under a prediction horizon of 720 and $\beta=4$. The reported results represent the average of five independent runs. As shown in the table, our TRACE method achieves the lowest parameter count (2.89M) among all parameter-efficient fine-tuning (PEFT) approaches for the backbone, demonstrating its superior parameter efficiency. While this comes at the cost of slightly longer training time compared to AdaLoRA, likely due to the overhead of the dynamic importance calculation in Gated DSIC, the inference time remains comparable. Furthermore, the reconstructed forecasting heads (e.g., Proj Down, Avg Pool) significantly reduce both the parameter count and the training/inference time compared to the standard linear probing head, highlighting their practical benefit in resource-constrained scenarios. 
\begin{table}[]
    \centering
    \caption{Computational cost on Electricity dataset.}
    \scalebox{0.8}{ % 0.9 表示缩放比例
    {\normalsize
    \begin{tabular}{lllll}
        \hline
         &Models & Parameters(M) &Trainging time(s) & Inference time (s) \\ \hline
        \multirow{5}{*}{\makecell{Forecasting \\Head}} &linear probing&35.38&387.32&41.70\\
        &\textbf{Proj Down}&8.99&251.34&27.45\\
        &\textbf{Avg Pool}&8.99&223.56&26.34\\
        &\textbf{Less Feature}&8.47&219.95&27.12\\
        &\textbf{Conv2D}&8.84&260.17&33.43\\ \hline
         \multirow{3}{*}{Backbone}&full LoRA&4.75&729.58&38.36\\
        &AdaLoRA&3.43&643.31&37.95\\
         &\textbf{TRACE}&2.89&942.86&38.17\\ \hline
       
    \end{tabular}}}
    \label{macs}
\end{table}

\subsubsection{Short-horizon Forecasting}
\begin{table}[H]
\centering
\caption{Zero-shot short-horizon forecasting performance on a subset of the M3 and M4 datasets measured using sMAPE.}
\scalebox{0.80}{ % 0.9 表示缩放比例
    {\normalsize
\begin{tabular}{cc|cc|cc|cc|cc}
\hline
\multicolumn{2}{c}{\multirow{2}{*}{Datasets}} & \multicolumn{2}{c}{$\;\;\mathrm{MOMENT}_{LP}\;\;$}    & \multicolumn{2}{c}{$\mathrm{MOMENT}_{LP+LoRA}$} & \multicolumn{2}{c}{$\mathrm{MOMENT}_{Proj\;down}$} & \multicolumn{2}{c}{\;\;\;\;\;TRACE\;\;\;\;\;} \\

\multicolumn{2}{c}{}& \multicolumn{1}{|c}{M4} & FR & \multicolumn{1}{|c}{M4}  & FR  & \multicolumn{1}{|c}{M4}     & FR     & \multicolumn{1}{|c}{M4}       & FR        \\ \hline

\multirow{3}{*}{M3}              & Yearly           &  17.03  & 17.13    &      17.21                    & 17.37     &  17.01                         & 17.85      &    16.88                          & 17.49                                      \\
                                    & Quarterly           & 10.47                       &  10.73   &   10.81                       &  11.36    &   10.34                          &10.74        & 10.25                              &   10.70                                         \\
                                    & Monthly           &16.35    &17.11    &16.49                         &17.26      & 16.32                            &   17.10     &  16.28                             & 17.03            \\ \hline
\multirow{3}{*}{M4}              & Yearly           & -                        &  15.11  &-                          &  15.37    & -                            & 14.98        &                   -            &    14.83                                       \\
                                    & Quarterly   &   -                      & 12.86    & -                         &13.14      & -                           &12.55         &     -                         &    12.03                                   \\
                                    & Monthly           &-                         & 16.26    & -                        &  16.73    & -                          &15.98        &  -                            &      15.87                                    \\ \hline

\
\end{tabular}}}
\label{short_exp}
\end{table}

% \begin{table}[H]
% \centering
% \scalebox{0.8}{ % 0.9 表示缩放比例
%     {\normalsize
% \begin{tabular}{l|c|cccc}
% \hline
% sMAPE&horizon&$\text{MOMENT}_{LP}$&$\text{MOMNET}_{LP+LoRA}$&$\text{MOMENT}_{Proj\;down}$&TRACE\\
% \hline
% M4-Yearly&6&49.625&48.371&47.717&45.362\\
% M4-Quarterly&8&57.476&58.115&55.371&54.984\\
% M4-Monthly&18&61.195&58.850&60.132&57.749\\
% \hline
% \end{tabular}}}
% \caption{Short-horizon forecasting performance on a subset of the M4 datasets measured using sMAPE.}
% \label{short_forecast}
% \end{table}
For short-horizon forecasting, we adopt the zero-shot setting proposed by Oreshkin \cite{shortyuce}. Specifically, we first fine-tune MOMENT on a \textbf{source dataset} with a forecasting head and then assess its performance on a \textbf{target dataset} without any additional fine-tuning. It is consistent with the setup in MOMENT \cite{moment}.

We evaluated our model's zero-forecasting performance on the M3 and M4 datasets. We maintained the same settings as in the MOMENT\cite{moment}, following the dataset splits for training, testing, and forecasting horizons as defined in the M3 and M4 competitions. The evaluation metric used was sMAPE (\cite{nbeats}; \cite{wu2022timesnet}). Due to the scarcity of daily, hourly, and weekly frequency data, it is challenging for deep models to fully leverage their zero-shot capabilities in these cases. Table \ref{source_target} presents the correspondence between source and target datasets. The setup for the Fred dataset follows \cite{moment} and \cite{shortyuce}.

Table \ref{short_exp} show that, under the sMAPE evaluation metric, the reconstructed forecasting head ($\mathrm{MOMENT}_{Proj\;down}$) and the TRACE method both outperform linear probing and full fine-tuning.

\begin{table}[]
    \centering
       \caption{Experimental settings for short-horizon forecasting experiments for varying source and target datasets.
}
    \scalebox{0.8}{ % 0.9 表示缩放比例
    {\normalsize
    \begin{tabular}{c|cc}
        \hline
       \makecell{Source Dataset$\to$\\ Target$\downarrow$} & M4 &Fred  \\ \hline
         \textbf{M4}& & \\
         Yearly & -&Yearly\\
         Quarterly & - &Yearly\\
         Monthly &-&Monthly\\
         \textbf{M3}&&\\
         Yearly& Yearly &Yearly \\
         Quarterly &Quarterly&Quarterly\\
         Monthly&Monthly &Monthly\\
         \hline
    \end{tabular}}}
    \label{source_target}
\end{table}

\subsubsection{Anomaly Detection}
\begin{table}[H]
\centering
\caption{Anomaly detection performance on selected datasets from the UCR Anomaly Archive}
\scalebox{0.8}{ % 0.9 表示缩放比例
    {\normalsize
\begin{tabular}{c|ccccccccc}
\hline
\multicolumn{1}{c|}{ \textbf{Adj-F1}} &
  \multicolumn{1}{c|}{MITDB} &
  \multicolumn{3}{c|}{ECG} &
  \multicolumn{3}{c|}{MGAB} &
  \multicolumn{2}{c}{SVDB} \\ \hline
\multicolumn{1}{c|}{Index} &
  \multicolumn{1}{c|}{100} &
  \multicolumn{1}{c|}{801} &
  \multicolumn{1}{c|}{803} &
  \multicolumn{1}{c|}{1406} &
  \multicolumn{1}{c|}{2} &
  \multicolumn{1}{c|}{3} &
  \multicolumn{1}{c|}{4} &
  \multicolumn{1}{c|}{842} &
  \multicolumn{1}{c}{859} \\ \hline
$\mathrm{MOMENT}_{LP}$      & 0.819  & 0.695 & 0.968 & 0.761 & 0.429 & 0.437 &0.459 & 0.828 & 0.864 \\
TRACE & 0.820 & 0.702 & 0.969 & 0.761 & 0.444 &0.448& 0.460 & 0.828 & 0.865\\
\hline
\end{tabular}}}
\label{anom}
\end{table}

In the anomaly detection task, we selected subsets of datasets from MITDB, ECG, MGAB, and SVDB. Since the task of anomaly detection involves a relatively small number of parameters in the forecasting head, which is not suitable for dimensionality reduction or reconstruction, in this context, TRACE is equivalent to Gated DSIC. Table \ref{anom} show that TRACE generally outperforms the linear probing.

\subsubsection{Natural Language Processing Task}
While our work focuses on continuous temporal data, evaluating on natural language tasks is necessary to assess the method’s applicability across sequential modalities. As natural language represents discrete sequences, such comparison reveals the adaptability and robustness of our approach beyond traditional time series, offering valuable insights into its generalization across both low- and high-frequency domains. These findings can serve as a reference and inspiration for future research exploring the intersection of temporal modeling and discrete sequence processing. 

We are currently conducting experiments on English datasets in the field of natural language processing. The datasets are sourced from the GLUE benchmark: paraphrase detection (MRPC, QQP), sentiment classification (SST-2), natural language inference (MNLI, RTE, QNLI), and linguistic acceptability (CoLA). We also consider three challenging QA datasets to evaluate the performance of our method on large language models: three commonsense question-answering benchmark tasks—ARC-e, ARC-c \cite{clark2018think}, and BoolQ \cite{clark2019boolq}.
% 我们现在在自然语言处理的英文数据集上进行实验。数据集来源于GLUE benchmark：paraphrase detection (MRPC, QQP), sentiment classification (SST-2), natural language inference (MNLI, RTE, QNLI), and linguistic acceptability (CoLA). 我们也考虑三个具有挑战性的QA数据集，用于评估我们的方法在large language model上的表现: three commonsense question-answering benchmark tasks—ARC-e, ARC-c \cite{clark2018think}, and BoolQ \cite{clark2019boolq}.
\begin{table*}[tb!]
\centering
\caption{\label{tab:nlp1} Overall comparison on the GLUE benchmark with the DeBERTaV3-base backbone. Results are averaged over five random experiments. Bold and underlined values indicate the best and second-best results, respectively. The initial LoRA rank is set to 4 and then halved according to our method. }
\resizebox{1.03\textwidth}{!}{
\renewcommand\arraystretch{1.2}
 \scalebox{1.2}{ % 0.9 表示缩放比例
    {\normalsize\begin{tabular}{ccccccccccc}
\hline
\textbf{Method}    &   \textbf{Trainable Params} & \textbf{Avg. Score} & \textbf{MNLI} & \textbf{QQP} & \textbf{QNLI} & \textbf{SST-2} & \textbf{CoLA} & \textbf{STS-B} & \textbf{MRPC} & \textbf{RTE} \\ 

\hline

Full Fine-tuning   &  184M  &   88.62    &   90.01    &  \textbf{91.10}    &   94.03    &   95.63    &   69.19    &   91.60    &   89.46    &   86.94   \\
\hline

Adapter   &    0.35M    &   87.57  &   90.28    &   89.41    &   93.52    &   94.38    &   67.80    &   91.08    &   89.75    &   84.36  \\

LoRA  &   0.33M    &   88.12  &   90.34    &   90.26    &   93.87    &   94.84    &   68.71    &   91.63    &   89.71    &   85.56  \\
AdaLoRA  &   0.35M    &   88.93  &   90.18    &   90.37    &   \underline{94.29}    &   95.62    &   70.04    &   91.57    &   \underline{90.34}    &   87.06  \\
MOELoRA    &   0.92M    &   88.89  &   90.26    &   90.31    &   94.08    &   95.53     &   70.47    &   91.52    &   90.26    &   \underline{87.64}  \\
DoRA   &   0.33M    &  \underline{ 89.00}  &  \underline{ 90.43}    &   90.16    &   94.17     &  \underline{ 95.80}    &   70.26    &   \underline{91.68}    &   90.12    &   87.38   \\
\hline
\multicolumn{11}{c}{\textbf{Proposed Method}}  \\
\hline

Our Method   &   0.33M  &    \textbf{89.51} &   \textbf{90.76}   &   \underline{90.66}    &   \textbf{94.72}   &   \textbf{96.17} &   \textbf{\underline{71.95}}  &   \textbf{92.33}  &   \textbf{90.86}   &   \textbf{88.63}  \\

\hline
\end{tabular}}}}
\end{table*}

\begin{table*}[tb!]
\centering
\caption{\label{tab:nlp2} Comparison of different PEFT methods for two backbone models (Qwen2.5 0.5B). Results are presented as accuracy for each task and averaged across five random experiments. The LLM used is the Qwen2.5 0.5B model, with the LoRA rank initially set to 32 and then halved.}
\resizebox{0.75\textwidth}{!}{
\begin{tabular}{c|c|c|c|c}
\hline
\textbf{Method} & \textbf{Trainable Params} &  \textbf{BoolQ} & \textbf{ARC-e} & \textbf{ARC-c}  \\ 
 \textbf{(Million)} & \textbf{(acc)} & \textbf{(acc)} & \textbf{(acc)} & \textbf{(acc)}  \\
\hline

\multicolumn{5}{c}{\textbf{Baseline(Qwen2.5 0.5B)}} \\
\hline

Adapter & 9.0M &  78.36 & 71.04 & 53.26  \\
LoRA & 8.8M &  78.94 & 72.78 & 54.38  \\
AdaLoRA & 8.9M &  80.32 & 73.96 & 54.23  \\

\hline

Our Method  & 8.8M &  82.49 & 74.67 & 55.64  \\

\hline
\end{tabular}}
\end{table*}
Experimental results on the GLUE benchmark are shown in Table \ref{tab:nlp1}, and those on the QA datasets are presented in Table \ref{tab:nlp2}. Our method, Gated-DSIC, achieves the best or second-best performance with the smallest number of trainable parameters in both settings.
% 在GLUE benchmark数据集上的实验结果如表\ref{tab:nlp1}所示；在QA datasets的结果如表\ref{tab:nlp2}所示。其中我们的方法，即Gated-DISC均在最小的训练参数的条件下，获得了最优或次优的表现性能。
\subsection{Ablation studies and further analysis}
In this section, we analyze the effects of dimensionality reduction factor $\beta$, forecasting head reconstruction methods, and LoRA rank on model performance. Results indicate that excessively large $\beta$ degrades accuracy, with optimal values typically being 4 or 8. For long-horizon forecasting, reconstruction method performance varies by task. Additionally, LoRA rank should remain moderate to avoid overfitting, as the pre-trained time-series backbone requires only minimal fine-tuning on the target dataset. Ablation studies confirm TRACE's strong stability.
\subsubsection{Impact of the Dimensionality Reduction Factor $\beta$}
\begin{figure}[H]
    \centering
    \includegraphics[width=0.83\linewidth]{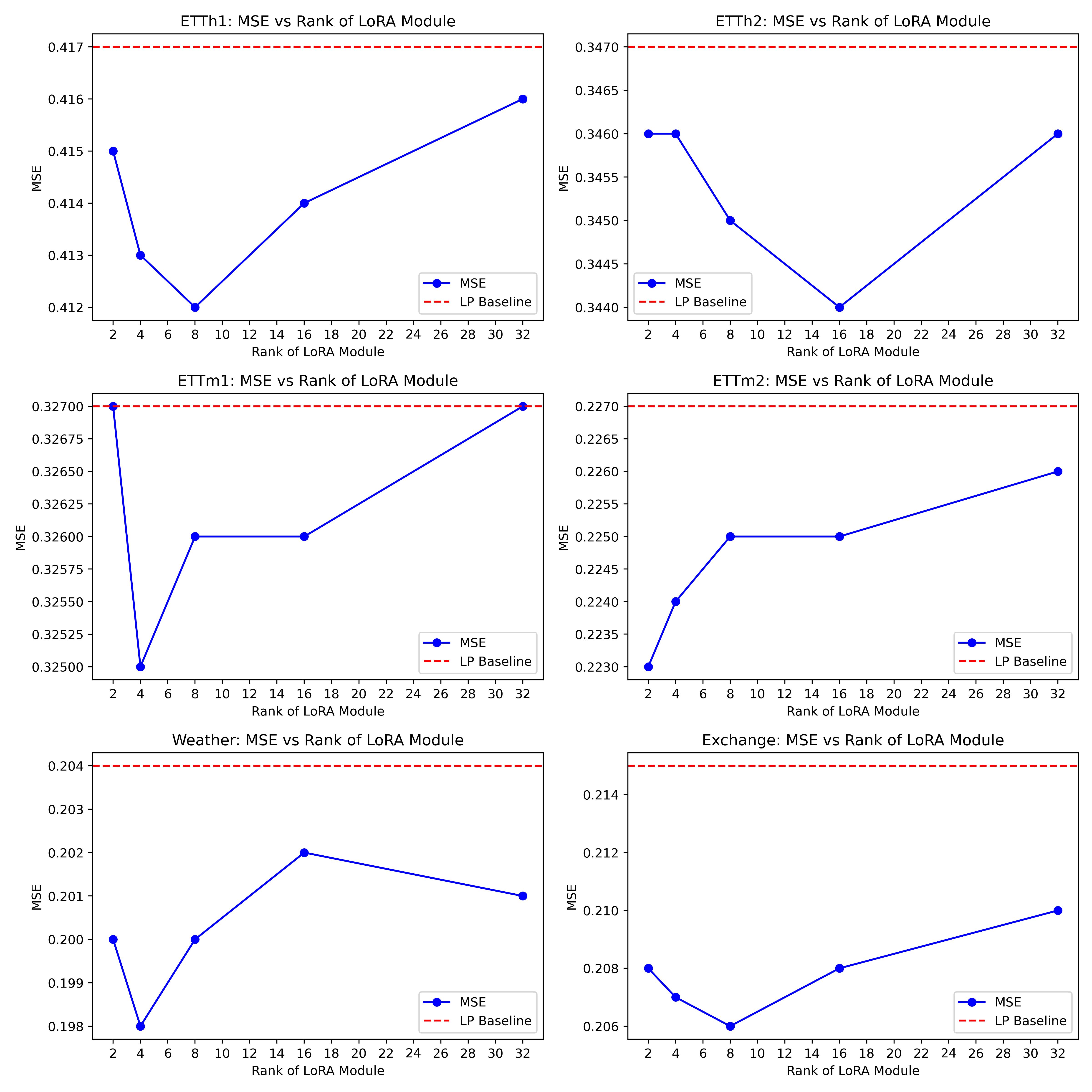}
    \caption{MSE Comparison between TRACE Method and Linear Probing on Different Datasets at Different $\beta$ (Prediction Horizon: 192, Forecasting Head: Projection Down)}
    \label{beta192}
\end{figure}
%图\ref{beta192}是TRACE方法在预测长度为192、重构预测头为Projection Down的情况下，在不同任务上降维因子的变化对于MSE的影响。分析结果可知，整体波动范围不大，在一个任务上的最大波动范围不超过1\%。通常情况下的最优降维因子$\beta$取4或者8，$\beta$取值过大会使得预测头可训练参数过小，从而降低模型预测性能。图\ref{beta720}为预测长度为720的结果，同样可以得到类似的结论。

\begin{figure}[htbp]
    \centering
    \includegraphics[width=0.83\linewidth]{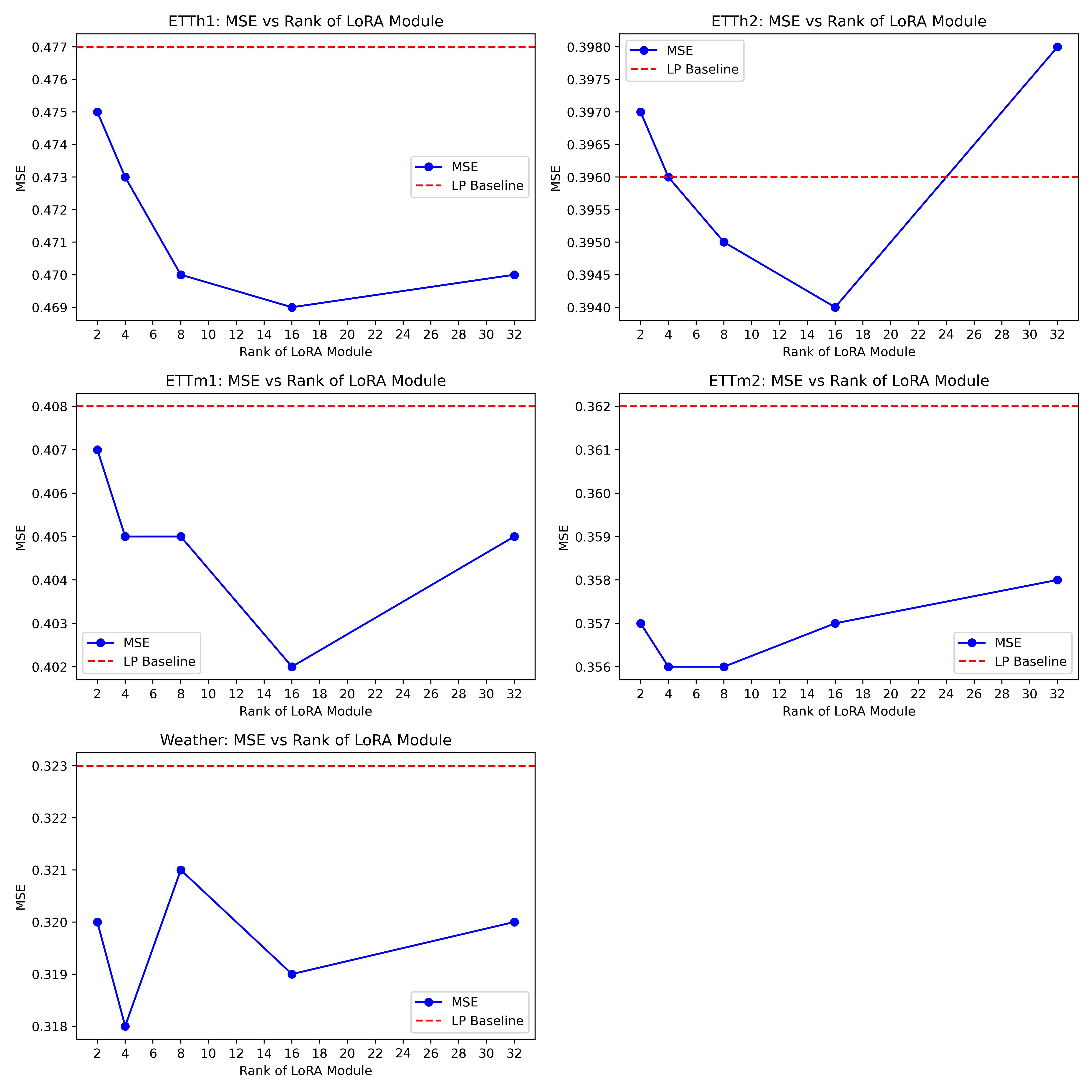}
    \caption{MSE Comparison between TRACE Method and Linear Probing on Different Datasets at Different $\beta$ (Prediction Horizon: 720, Forecasting Head: Projection Down)}
    \label{beta720}
\end{figure}

Figure \ref{beta192} illustrates the effect of varying the dimensionality reduction factor on MSE under the TRACE method, with a forecast horizon of 192 and the Projection Down reconstruction for different tasks. The analysis shows that the overall fluctuation is minimal, with the maximum fluctuation in a single task not exceeding 1\%. The optimal dimensionality reduction factor, $\beta$, typically takes values of 4 or 8. A larger value of $\beta$ reduces the number of trainable parameters in the forecasting head, which in turn negatively impacts the model's forecasting performance. Figure \ref{beta720} presents the results for a forecast horizon of 720, where similar conclusions can be drawn.
\subsubsection{Impact of Different Forecasting Heads}

\begin{table}[H]
\centering
\caption{Long-term forecasting performance measured using Mean Squared Error (MSE) and Mean Absolute Error (MAE) in different reconstruction forecasting head. Bold values indicate the best.}
\scalebox{0.7}{ % 0.9 表示缩放比例
    {\normalsize
\begin{tabular}{cc|cc|cc|cc|cc|cc}
\hline
\multicolumn{2}{c}{\multirow{2}{*}{ \makecell{Forecasting\\Horizon}}} & \multicolumn{2}{c}{$\;\;\mathrm{MOMENT}_{LP}\;\;$}    & \multicolumn{2}{c}{$\mathrm{TRACE}_{Less\;Feature}$} & \multicolumn{2}{c}{$\mathrm{TRACE}_{Avg\;Pool}$}&\multicolumn{2}{c}{$\mathrm{TRACE}_{Conv2D}$} & \multicolumn{2}{c}{$\mathrm{TRACE}_{Proj\;Down}$}  \\

\multicolumn{2}{c}{}& \multicolumn{1}{|c}{MSE} & MAE & \multicolumn{1}{|c}{MSE}  & MAE  & \multicolumn{1}{|c}{MSE}     & MAE     & \multicolumn{1}{|c}{MSE}       & MAE   & \multicolumn{1}{|c}{MSE}       & MAE          \\ \hline

\multirow{3}{*}{ETTh1}              & 192           &  0.417  & 0.430    &      0.413                    & 0.428     &  0.414                           & 0.429       &0.423 &0.435 &    \textbf{0.412}                           & \textbf{0.427}                                      \\
                                    & 336           & 0.432                     &  0.449   &   0.428                       &  0.445    &   \textbf{0.427}                          &\textbf{0.443}      & 0.437&0.455    & 0.428                              &   0.444                                          \\
                                    & 720           &0.476    &0.487     &\textbf{0.469}                          &\textbf{0.480}      & 0.471                            &   0.482     &0.479&0.491  &  0.470                             & 0.481            \\ \hline
\multirow{3}{*}{ETTh2}              & 192           & 0.347                        &  0.389   &0.347                          &  0.388    & \textbf{0.345}                            & 0.386        &   0.353   &0.394&              \textbf{0.345}            &    \textbf{0.385}                                       \\
                                    & 336   &   0.374                      & 0.413    & 0.369                         &\textbf{0.407}      & \textbf{0.368}                            &\textbf{0.407}         &   0.377  &0.417&  0.370                         & 0.409                                      \\
                                    & 720           &0.396                         & 0.435    &  \textbf{0.395}                        &  \textbf{0.434}    &  0.396                           &0.435       &0.416& 0.457  &  \textbf{0.395}                             &      \textbf{0.434}                                    \\ \hline
\multirow{3}{*}{ETTm1}              & 192           &   0.327                      &  0.368   &0.326                          & 0.367     & \textbf{0.325}                            & \textbf{0.365}       &0.333& 0.375&       0.326                        & 0.366                                       \\
                                    & 336           &0.355                         & 0.388    &   \textbf{0.351}                       & 0.385     &0.352                             & 0.386        &       0.354 &0.388&    \textbf{0.351}                    &       \textbf{ 0.384}                                  \\
                                    & 720           & 0.408                        &  0.419   &   \textbf{0.403}                       & \textbf{0.414}     &         0.404                    & 0.415        &   0.417&0.428&0.405                             &           0.416                               \\ \hline
\multirow{3}{*}{ETTm2}              & 192           &    0.227                     &  0.299   &                    0.228     &  0.300    & \textbf{0.226}                            &   \textbf{0.298}      &    0.231   &0.305&               0.228          &    0.299                                    \\
                                    & 336           & 0.276                        &  0.329   &  0.276                        & 0.328     &   \textbf{0.273}                          &\textbf{0.325}         &   0.287 &0.339&  0.274                          &                        0.325                \\
                                    & 720           &  0.360                       & 0.386    &  0.358                        & 0.384     &   \textbf{0.354}                          &  \textbf{0.380}       &  0.373 &0.401& 0.359                             & 0.383                                      \\ \hline
\multirow{3}{*}{Weather}            & 192           &     0.204                    &0.254     &                   0.201       & 0.248     &                    \textbf{0.200}         & \textbf{0.248}        &                    0.211 &0.260&     \textbf{0.200}      &     \textbf{0.248}                                     \\
                                    & 336           &0.253                         & 0.294    &\textbf{0.250}                          & \textbf{0.291}     &          0.251                   &   0.293      &  0.267&0.308&0.251                              &  0.292                                     \\
                                    & 720           &   0.323                      &  0.340   &  0.319                        & 0.339     & \textbf{0.317}                            & \textbf{0.336}        &   0.328&0.345&\textbf{0.317}                             &    0.337                                     \\ \hline
\multirow{3}{*}{Exchange}           & 96            &    0.122                     &  0.249   &               \textbf{0.103}           &  \textbf{0.228}    & 0.105                            &   0.229      &                   0.119        &0.245&  0.108   &     0.233                                     \\
                                    & 192           & 0.225                        &  0.342   & 0.211                         &0.330      &  0.208                          & 0.327       &  0.228  &0.345&  \textbf{0.206}                           &    \textbf{0.326}                                    \\
                                    & 336           &     0.317                    &0.434     &  0.314                        &  0.432    &           \textbf{0.312}                  &    \textbf{0.430}     &   0.329  &0.347&       0.313                    &             \textbf{0.430}                            \\
                                    \hline
\end{tabular}}}
\label{diff_head}
\end{table}

Under the condition of the same dimensionality reduction factor ($\beta=8$), we compared the performance of the TRACE method with different reconstruction approaches for the forecasting head and the original linear probing method on long-term forecasting datasets. Specifically, $\mathrm{TRACE}_{Proj\;Down}$, $\mathrm{TRACE}_{Less\;Feature}$, $\mathrm{TRACE}_{Avg\;Pool}$, and $\mathrm{TRACE}_{Conv2D}$ refer to the different forecasting head reconstruction methods previously introduced. 

The comparison results in Table \ref{diff_head}, in nearly all tasks, the TRACE method outperforms Linear Probing or performs equivalently. However, when reconstructing the forecasting head with Conv2D, the performance is generally suboptimal, and in most cases, it is inferior to Linear Probing, with a performance gap of within 3.8\%. A possible reason for this is that 2D convolution may not be well-suited for extracting features from time series with independent channels, as fluctuations between covariates can introduce interference.

% 但重构预测头为Conv2D时效果一般，大多数部分情况下效果不如Linear Probing，差距在3.8\%以内，分析其原因可能为二维卷积不适用于通道独立的时间序列提取特征，协变量之间的波动会造成干扰。
However, the performance of different reconstruction methods varies across tasks, with no single method being consistently optimal for all tasks. This suggests that, in practical applications, it may be beneficial to tailor the forecasting head to suit the specific task at hand.

\subsubsection{Impact of LoRA Rank}
\begin{figure}[H]
    \centering
    \includegraphics[width=1\linewidth]{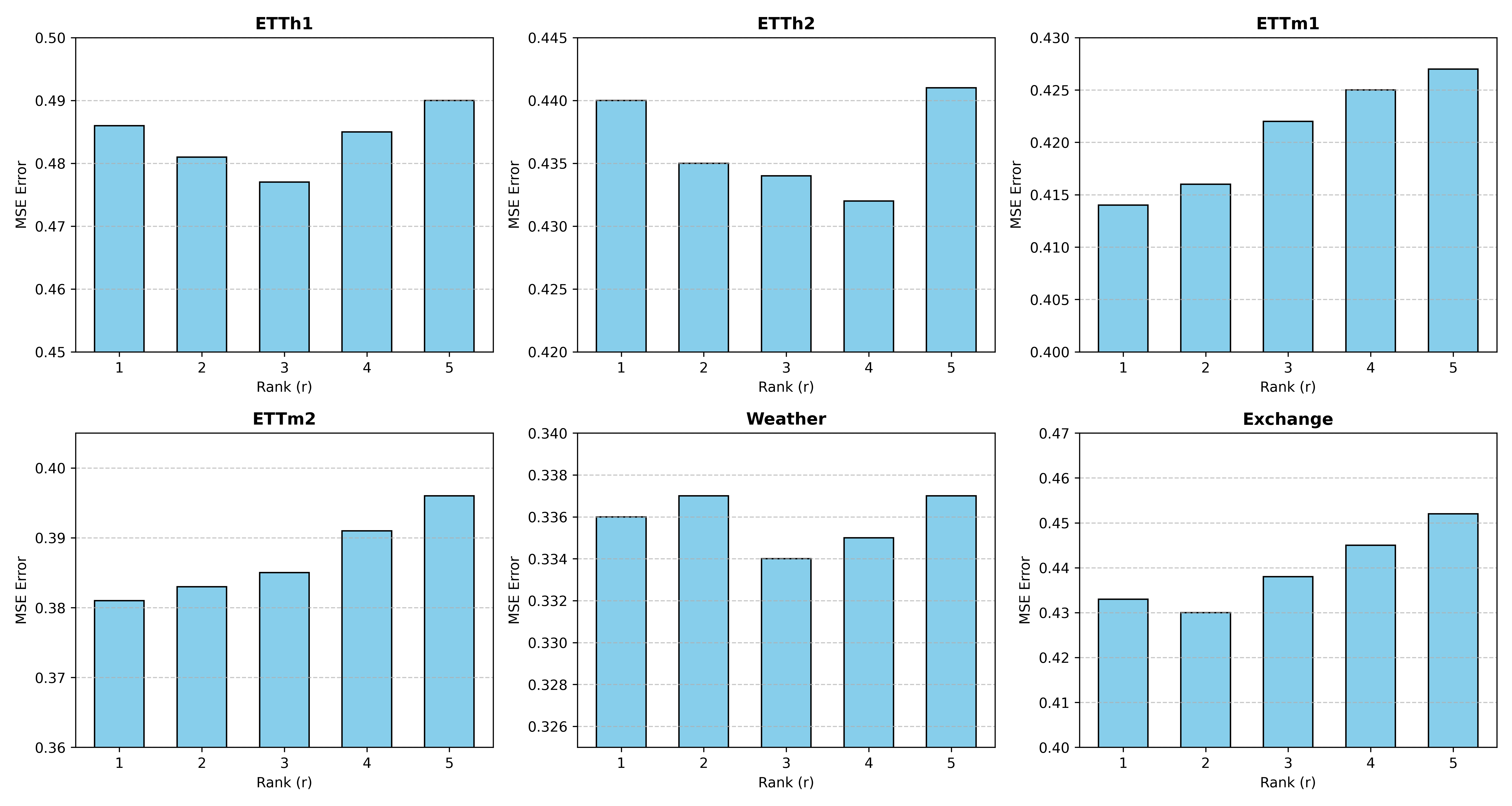}
    \caption{The impact of the rank size of the incremental matrix on model performance measured using Mean Squared Error (MSE)}
    \label{imp_lora}
\end{figure}
% 图$\ref{imp_lora}$展示了仅改变LoRA的秩 $\gamma$ 而其余参数保持和主实验设置相同的条件下，TRACE方法在不同长时预测数据集上的变化。通过分析可知，秩的最优取值在2$\sim$3。该消融实验也证明了，时序基座大模型在进行LoRA微调时，秩$\gamma$不适宜取值太大。
Figure $\ref{imp_lora}$ illustrates the performance variations of the TRACE method on different long-term forecasting datasets when only the LoRA rank $\gamma$ is changed, while keeping all other parameters consistent with the main experiment settings. The analysis indicates that the optimal rank value falls within the range of 2 to 3. This ablation study also confirms that when fine-tuning time series foundation models with LoRA, selecting an excessively large rank $\gamma$ is not advisable.

\subsubsection{Case Analysis}
\begin{figure}[H]
    \centering
    \includegraphics[width=1.1\linewidth]{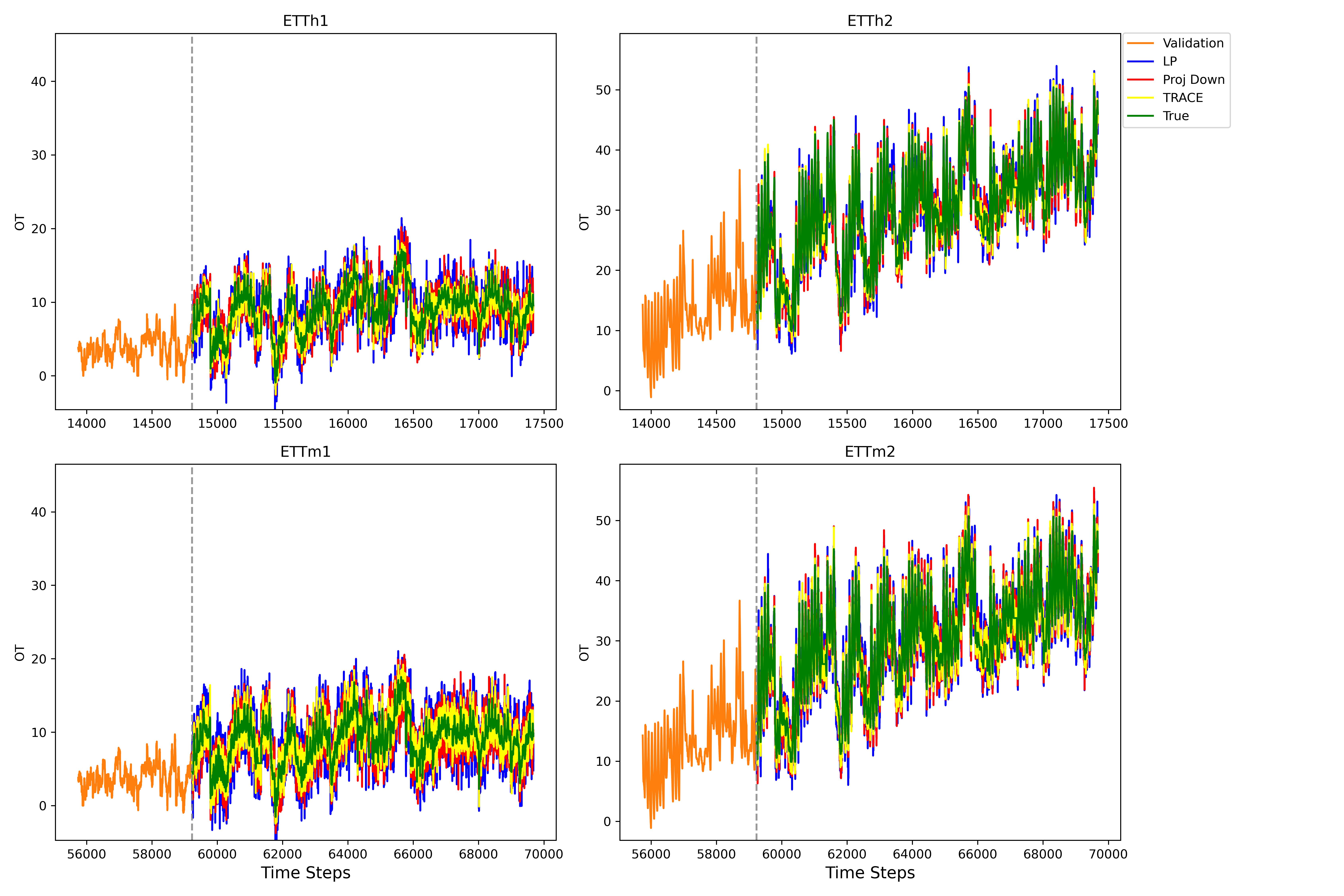}
    \caption{
    Comparison of true values on the ETT long-horizon forecasting dataset's test set with predicted values from different models
% ETT长时预测数据集在测试集上的真实值以及不同模型设置下的预测值对比
    }
    \label{case_analysis}
\end{figure}

% 我们分别对ETT的四个数据集(ETTh1,ETTh2,ETTm1,ETTm2)进行案例分析，将测试集的真实值时间序列，$\mathrm{MOMENT}_{LP}$，$\mathrm{MOMENT}_{Proj\;down}$以及$\mathrm{TRACE}_{Proj\;Down}$的预测值序列画出来如图\ref{case_analysis}所示。可以看出TRACE方法的预测值更加贴近真实值，在ETTh1和ETTm1数据上表现更为明显。
We conducted a case analysis on four ETT datasets (ETTh1, ETTh2, ETTm1, and ETTm2) by plotting the ground truth time series of the test set along with the predicted sequences from $\mathrm{MOMENT}_{LP}$, $\mathrm{MOMENT}_{Proj\;down}$, and $\mathrm{TRACE}_{Proj\;Down}$, as illustrated in Figure \ref{case_analysis}. In the figure, the predictions generated by the TRACE method are represented by the yellow curve, which more closely aligns with the green curve denoting the ground truth compared to the results from other models. 

\begin{figure}[htbp]
    \centering
    \begin{subfigure}[b]{0.25\linewidth}
        \includegraphics[width=\linewidth]{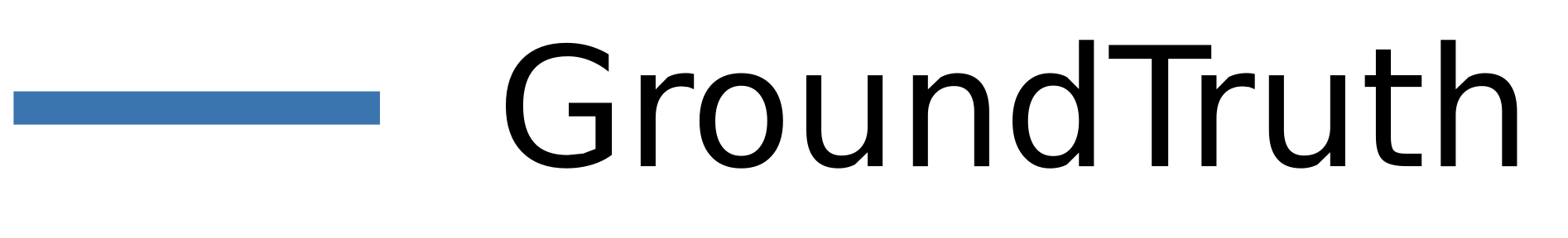}
    \end{subfigure}
    % \hfill
    \hspace{0.23\linewidth}
    \centering
    \begin{subfigure}[b]{0.25\linewidth}
        \includegraphics[width=\linewidth]{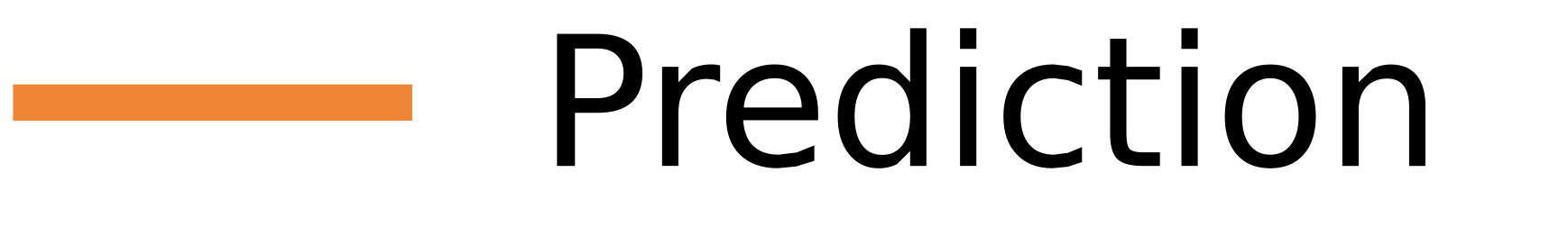}
    \end{subfigure}
    \begin{subfigure}[b]{0.48\linewidth}
        \includegraphics[width=\linewidth]{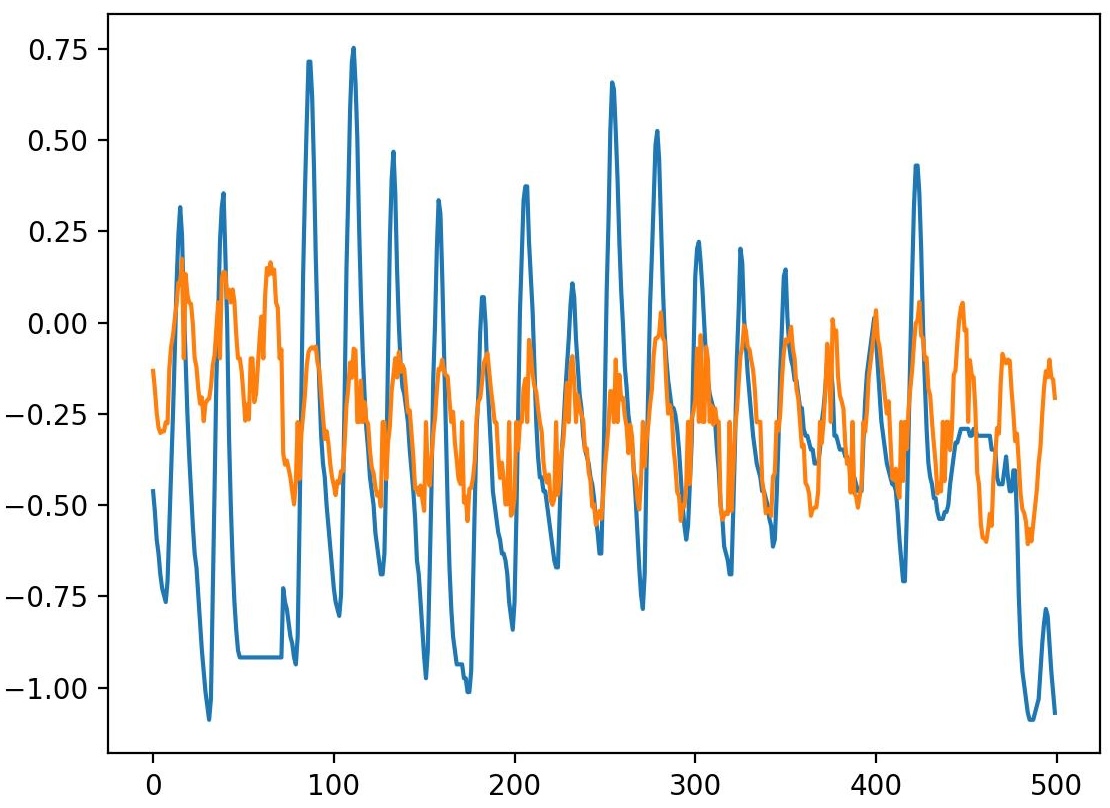}
        \caption{$\mathrm{MOMENT}_{LP}$}
        \label{c1:img1}
    \end{subfigure}
    % \hfill
    \hspace{0.001\linewidth}
    \begin{subfigure}[b]{0.48\linewidth}
        \includegraphics[width=\linewidth]{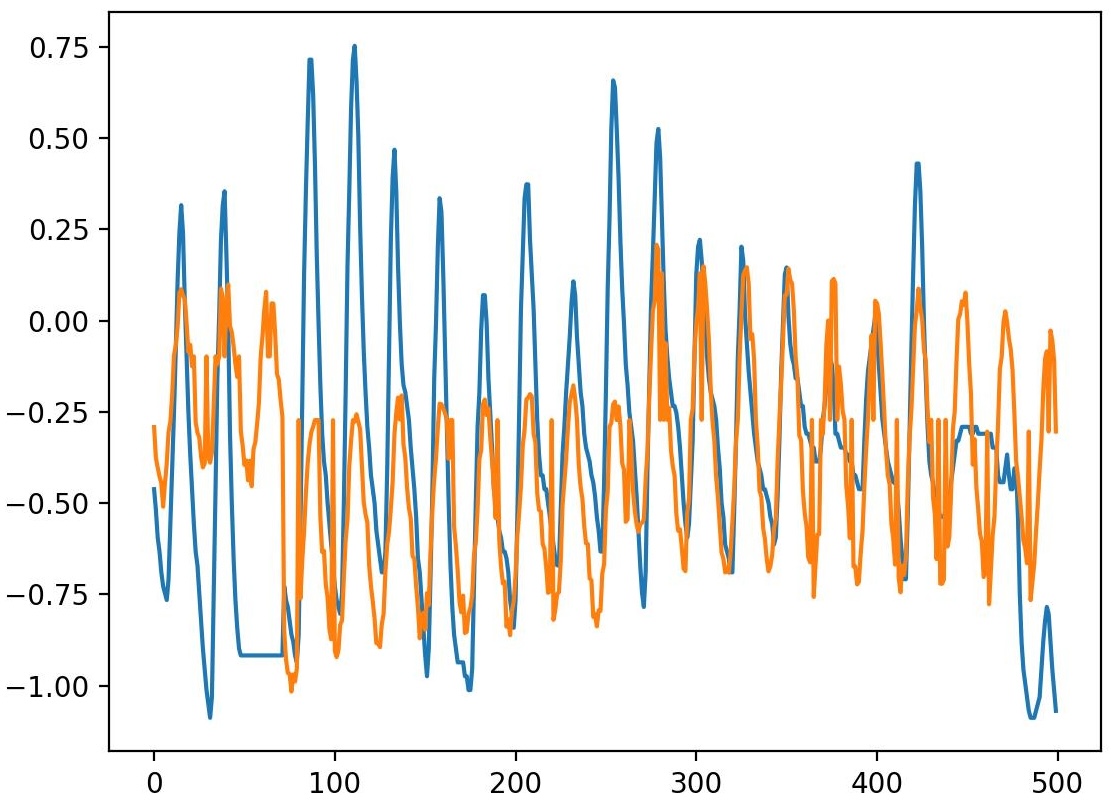}
        \caption{$\mathrm{MOMENT}_{Proj\;Down}$}
        \label{c1:img2}
    \end{subfigure}
    \vspace{1em} % 增加行间距
    \begin{subfigure}[b]{0.48\linewidth}
        \includegraphics[width=\linewidth]{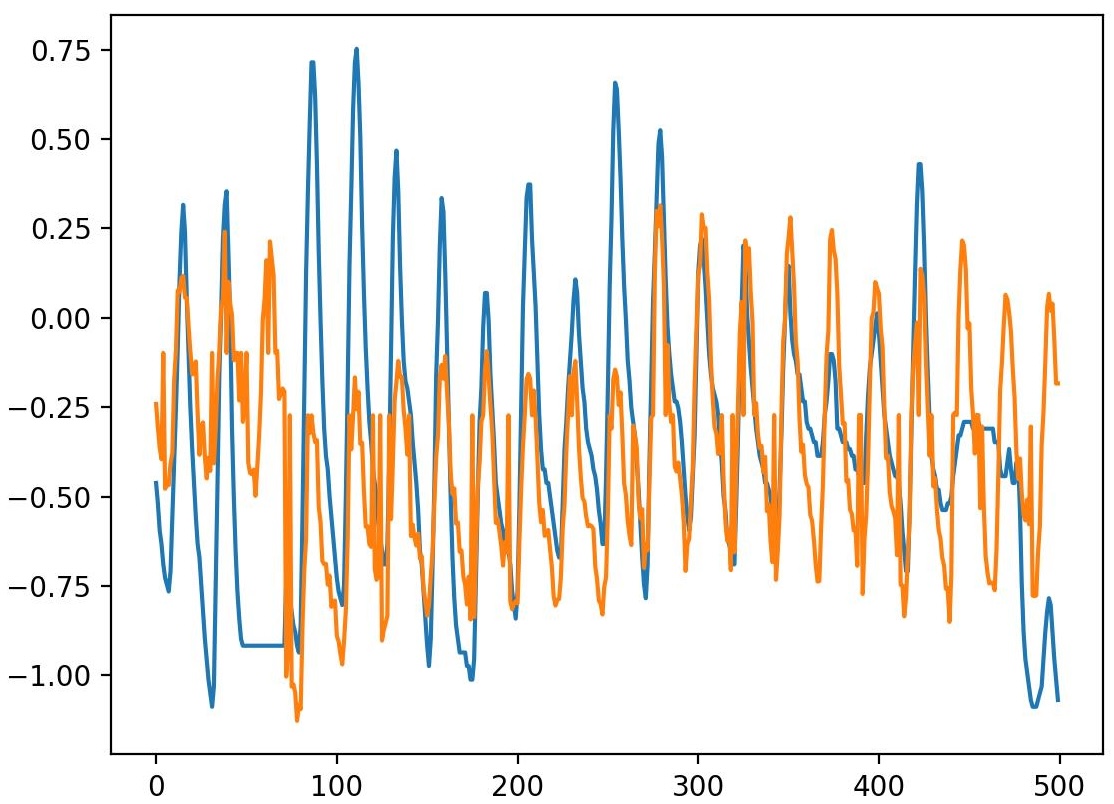}
        \caption{$\mathrm{MOMENT}_{Avg\;Pool}$}
        \label{c1:img3}
    \end{subfigure}
    % \hfill
    \hspace{0.001\linewidth}
    \begin{subfigure}[b]{0.48\linewidth}
        \includegraphics[width=\linewidth]{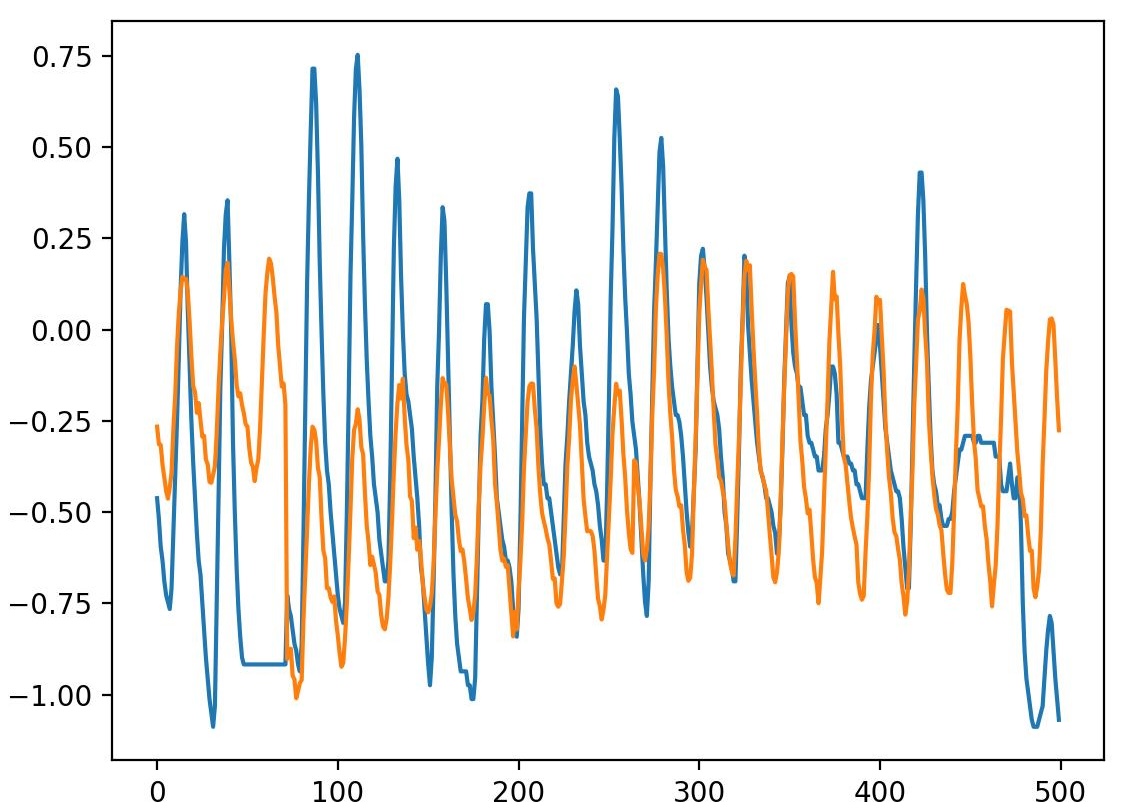}
        \caption{$\mathrm{MOMENT}_{Less\;Feature}$}
        \label{c1:img4}
    \end{subfigure}
    \vspace{1em}
    \begin{subfigure}[b]{0.48\linewidth}
        \includegraphics[width=\linewidth]{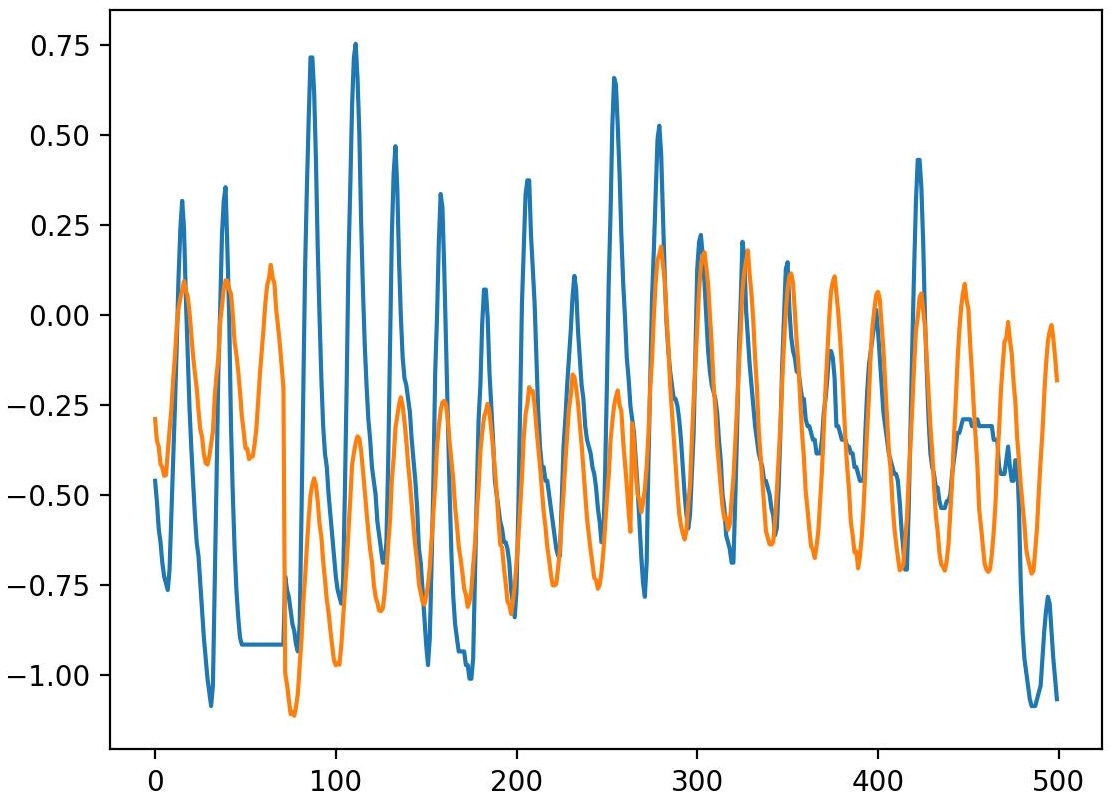}
        \caption{$\mathrm{MOMENT}_{AdaLoRA+LP}$}
        \label{c1:img5}
    \end{subfigure}
    \hspace{0.001\linewidth}
    \begin{subfigure}[b]{0.48\linewidth}
        \includegraphics[width=\linewidth]{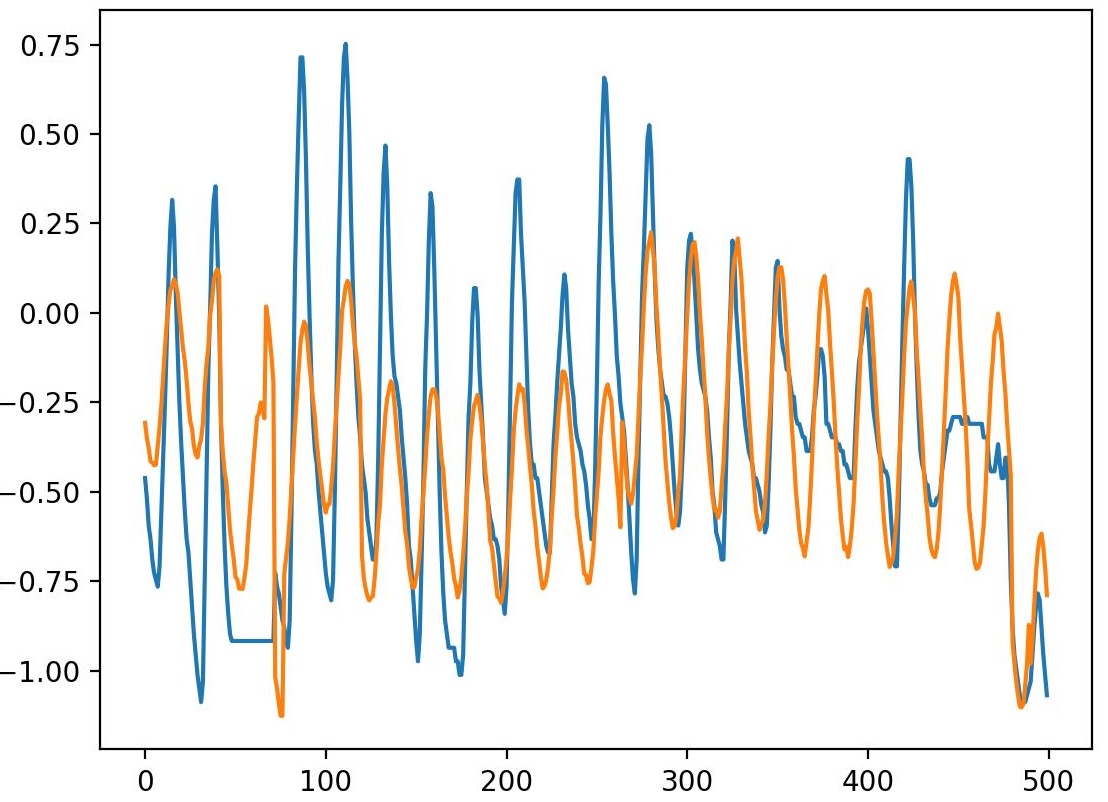}
        \caption{$\mathrm{TRACE}_{Proj\;Down}$}
        \label{c1:img6}
    \end{subfigure}
    
    \caption{Forecasting cases using the input-512-predict-192 configuration.}
    \label{fig:allimages192}
\end{figure}

\begin{figure}[htbp]
    \centering
    \begin{subfigure}[b]{0.25\linewidth}
        \includegraphics[width=\linewidth]{true.png}
    \end{subfigure}
    % \hfill
    \hspace{0.23\linewidth}
    \centering
    \begin{subfigure}[b]{0.25\linewidth}
        \includegraphics[width=\linewidth]{pred.png}
    \end{subfigure}
    \centering
    \begin{subfigure}[b]{0.48\linewidth}
        \includegraphics[width=\linewidth]{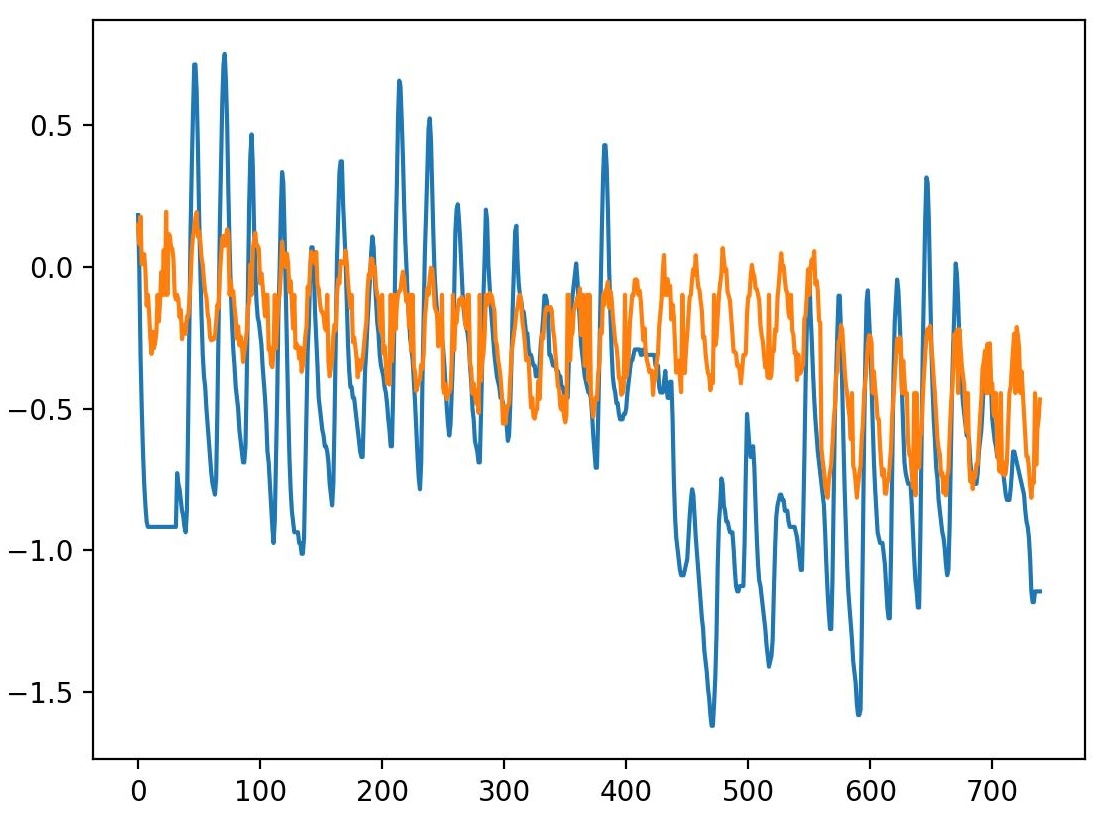}
        \caption{$\mathrm{MOMENT}_{LP}$}
        \label{c2:img1}
    \end{subfigure}
    % \hfill
    \hspace{0.001\linewidth}
    \begin{subfigure}[b]{0.48\linewidth}
        \includegraphics[width=\linewidth]{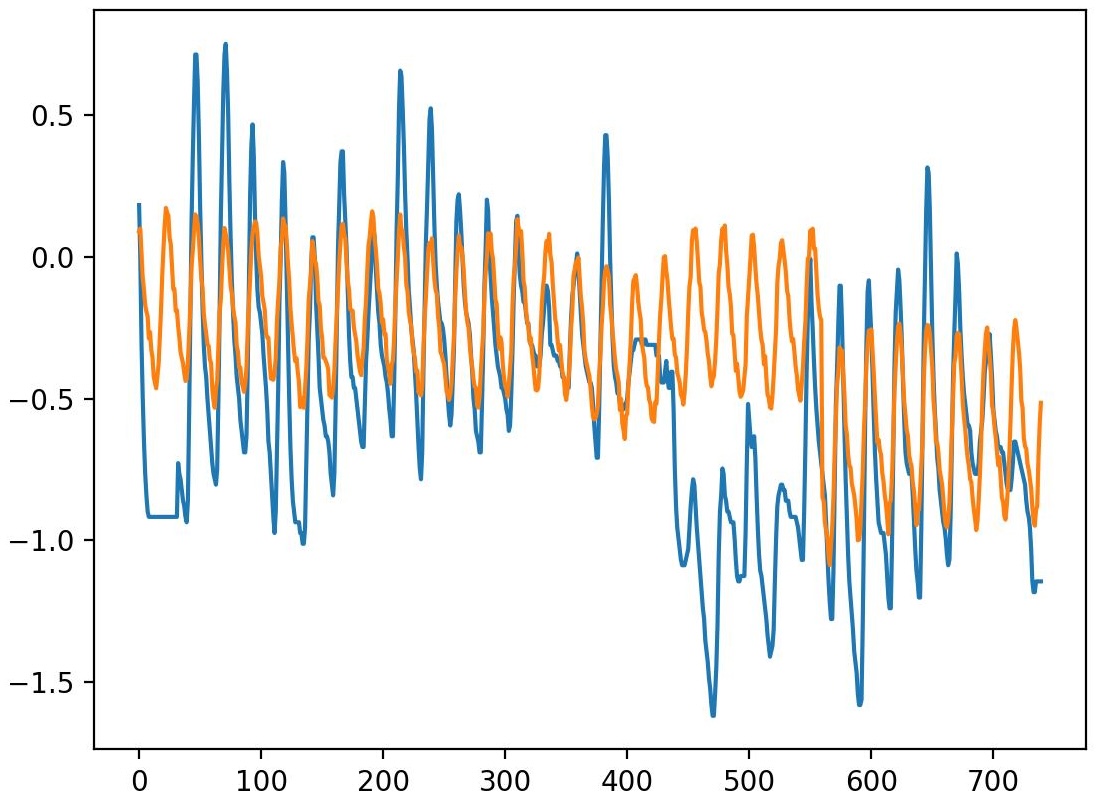}
        \caption{$\mathrm{MOMENT}_{Proj\;Down}$}
        \label{c2:img2}
    \end{subfigure}
    \vspace{1em} % 增加行间距
    \begin{subfigure}[b]{0.48\linewidth}
        \includegraphics[width=\linewidth]{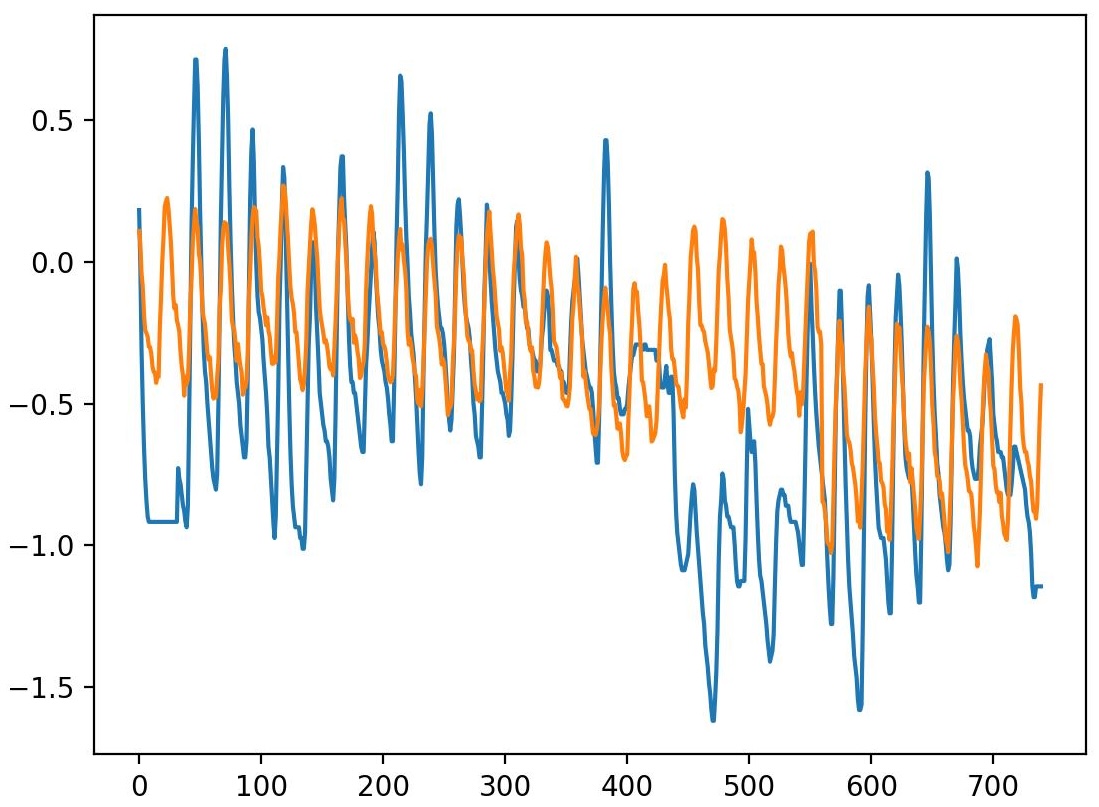}
        \caption{$\mathrm{MOMENT}_{Avg\;Pool}$}
        \label{c2:img3}
    \end{subfigure}
    % \hfill
    \hspace{0.001\linewidth}
    \begin{subfigure}[b]{0.48\linewidth}
        \includegraphics[width=\linewidth]{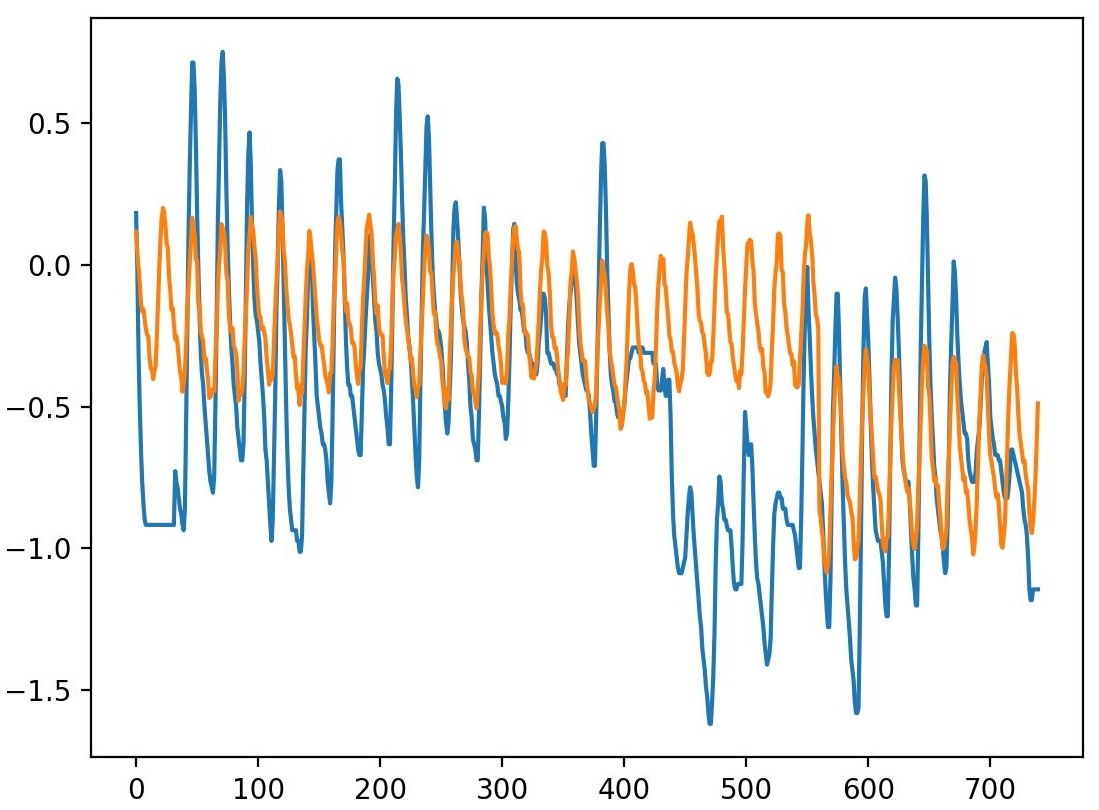}
        \caption{$\mathrm{MOMENT}_{Less\;Feature}$}
        \label{c2:img4}
    \end{subfigure}
    \vspace{1em}
    \begin{subfigure}[b]{0.48\linewidth}
        \includegraphics[width=\linewidth]{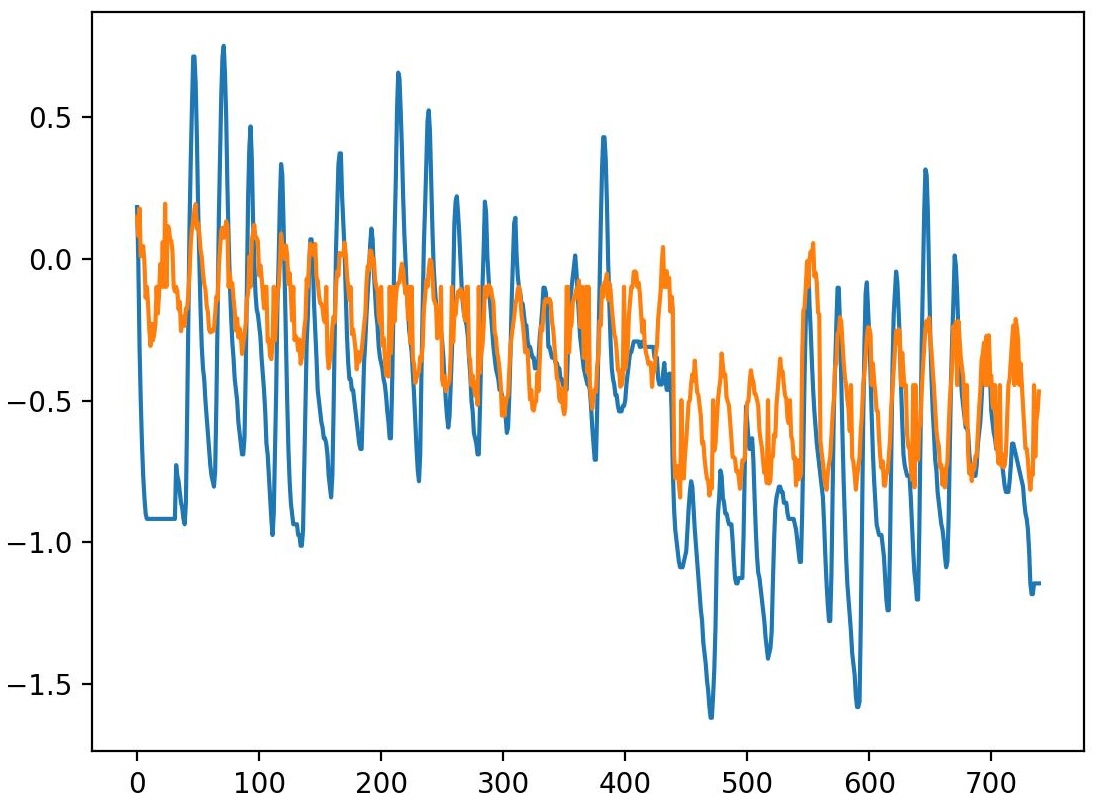}
        \caption{$\mathrm{MOMENT}_{AdaLoRA+LP}$}
        \label{c2:img5}
    \end{subfigure}
    \hspace{0.001\linewidth}
    \begin{subfigure}[b]{0.48\linewidth}
        \includegraphics[width=\linewidth]{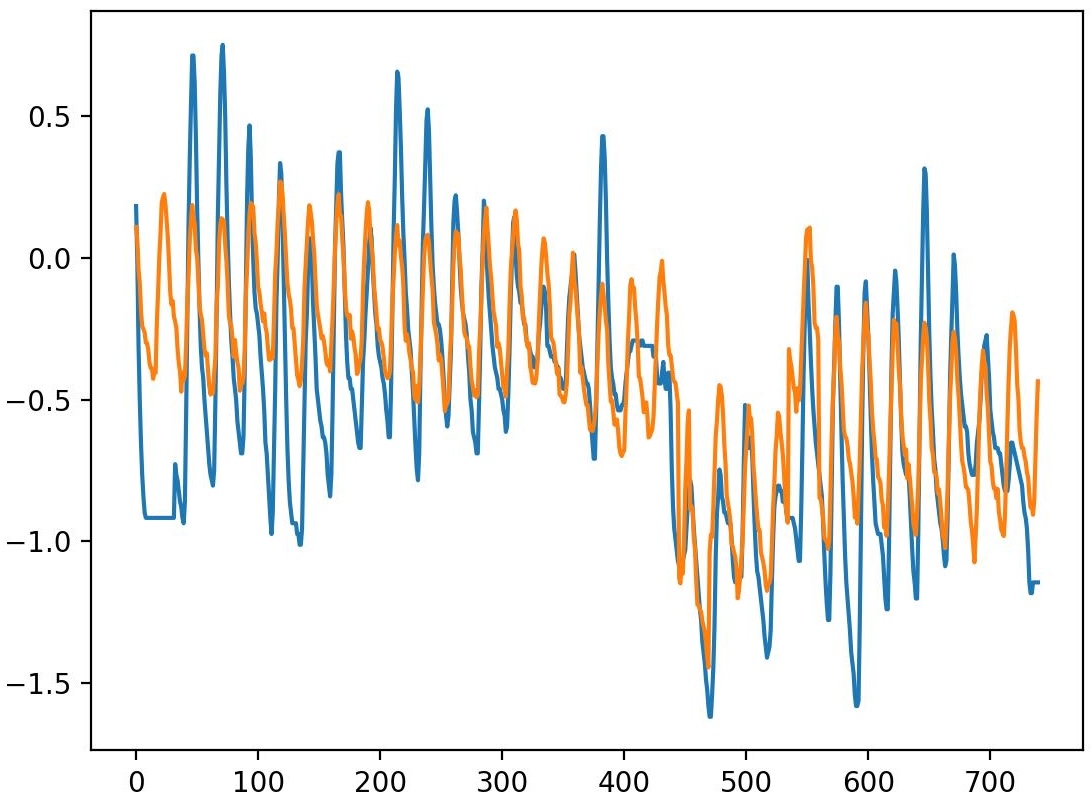}
        \caption{$\mathrm{TRACE}_{Proj\;Down}$}
        \label{c2:img6}
    \end{subfigure}
    
    \caption{Forecasting cases using the input-512-predict-720 configuration.}
    \label{fig:allimages720}
\end{figure}

To provide a more intuitive comparison, we visualize partial forecasts on the ETTh2 test set. For a prediction horizon of 192 (Figure \ref{fig:allimages192}, TRACE in Figure \ref{c1:img6}), our method shows better alignment with the true values, especially during volatile periods (e.g., steps 460–500). With a horizon of 720 (Figure \ref{fig:allimages720}), TRACE in Figure \ref{c2:img6}  captures the drop trend around step 500 more accurately than competing methods.

\subsection{Conclusions and future work}
In this work, we presented TRACE, a parameter-efficient fine-tuning framework tailored for time series foundation models. Our approach addresses two critical bottlenecks: the parameter inefficiency of standard forecasting heads and the biased importance evaluation in existing PEFT methods. Through extensive experiments, we demonstrated that TRACE’s reconstructed heads and the Gated DSIC mechanism consistently yield superior performance across a range of tasks.

A current limitation is that the optimal set of LoRA modules may vary across tasks. For future work, we plan to investigate task-agnostic or meta-learning-based strategies to identify a universal set of adaptable modules. Furthermore, we aim to explore the integration of TRACE with multimodal time series (e.g., combining sensor data with textual logs) to unlock new applications in real-world scenarios.

\newpage

\bibliographystyle{elsarticle-num} 
\bibliography{reference_new.bib}

@article{godahewa2021monash,
  title={Monash time series forecasting archive},
  author={Godahewa, Rakshitha and Bergmeir, Christoph and Webb, Geoffrey I and Hyndman, Rob J and Montero-Manso, Pablo},
  journal={arXiv preprint arXiv:2105.06643},
  year={2021}
}

@article{shazeer2020glu,
  title={Glu variants improve transformer},
  author={Shazeer, Noam},
  journal={arXiv preprint arXiv:2002.05202},
  year={2020}
}

@inproceedings{shortyuce,
  title={Meta-learning framework with applications to zero-shot time-series forecasting},
  author={Oreshkin, Boris N and Carpov, Dmitri and Chapados, Nicolas and Bengio, Yoshua},
  booktitle={Proceedings of the AAAI conference on artificial intelligence},
  volume={35},
  number={10},
  pages={9242--9250},
  year={2021}
}

@article{jiang2025tcm,
  title={TCM: An efficient lightweight MLP-based network with affine transformation for long-term time series forecasting},
  author={Jiang, Hongwei and Liu, Dongsheng and Ding, Xinyi and Chen, Yaning and Li, Hongtao},
  journal={Neurocomputing},
  volume={617},
  pages={128960},
  year={2025},
  publisher={Elsevier}
}

@inproceedings{etth,
  author    = {Haoyi Zhou and
               Shanghang Zhang and
               Jieqi Peng and
               Shuai Zhang and
               Jianxin Li and
               Hui Xiong and
               Wancai Zhang},
  title     = {Informer: Beyond Efficient Transformer for Long Sequence Time-Series Forecasting},
  booktitle = {The Thirty-Fifth {AAAI} Conference on Artificial Intelligence, {AAAI} 2021, Virtual Conference},
  volume    = {35},
  number    = {12},
  pages     = {11106--11115},
  publisher = {{AAAI} Press},
  year      = {2021},
}

@article{hyndman2002state,
  title={A state space framework for automatic forecasting using exponential smoothing methods},
  author={Hyndman, Rob J and Koehler, Anne B and Snyder, Ralph D and Grose, Simone},
  journal={International Journal of forecasting},
  volume={18},
  number={3},
  pages={439--454},
  year={2002},
  publisher={Elsevier}
}

@article{clark2018think,
  title={Think you have solved question answering? try arc, the ai2 reasoning challenge},
  author={Clark, Peter and Cowhey, Isaac and Etzioni, Oren and Khot, Tushar and Sabharwal, Ashish and Schoenick, Carissa and Tafjord, Oyvind},
  journal={arXiv preprint arXiv:1803.05457},
  year={2018}
}

@article{clark2019boolq,
  title={Boolq: Exploring the surprising difficulty of natural yes/no questions},
  author={Clark, Christopher and Lee, Kenton and Chang, Ming-Wei and Kwiatkowski, Tom and Collins, Michael and Toutanova, Kristina},
  journal={arXiv preprint arXiv:1905.10044},
  year={2019}
}

@article{var,
  title={Vector autoregressions},
  author={Stock, James H and Watson, Mark W},
  journal={Journal of Economic perspectives},
  volume={15},
  number={4},
  pages={101--115},
  year={2001},
  publisher={American Economic Association}
}

@inproceedings{arima,
  title={Stock price prediction using the ARIMA model},
  author={Ariyo, Adebiyi A and Adewumi, Adewumi O and Ayo, Charles K},
  booktitle={2014 UKSim-AMSS 16th international conference on computer modelling and simulation},
  pages={106--112},
  year={2014},
  organization={IEEE}
}

@inproceedings{chen2016xgboost,
  title={Xgboost: A scalable tree boosting system},
  author={Chen, Tianqi and Guestrin, Carlos},
  booktitle={Proceedings of the 22nd acm sigkdd international conference on knowledge discovery and data mining},
  pages={785--794},
  year={2016}
}

@article{ke2017lightgbm,
  title={Lightgbm: A highly efficient gradient boosting decision tree},
  author={Ke, Guolin and Meng, Qi and Finley, Thomas and Wang, Taifeng and Chen, Wei and Ma, Weidong and Ye, Qiwei and Liu, Tie-Yan},
  journal={Advances in neural information processing systems},
  volume={30},
  year={2017}
}

@article{gru,
  title={Empirical evaluation of gated recurrent neural networks on sequence modeling},
  author={Chung, Junyoung and Gulcehre, Caglar and Cho, KyungHyun and Bengio, Yoshua},
  journal={arXiv preprint arXiv:1412.3555},
  year={2014}
}

@article{nbeats,
  title={N-BEATS: Neural basis expansion analysis for interpretable time series forecasting},
  author={Oreshkin, Boris N and Carpov, Dmitri and Chapados, Nicolas and Bengio, Yoshua},
  journal={arXiv preprint arXiv:1905.10437},
  year={2019}
}

@inproceedings{tcn,
  title={Temporal convolutional networks for action segmentation and detection},
  author={Lea, Colin and Flynn, Michael D and Vidal, Rene and Reiter, Austin and Hager, Gregory D},
  booktitle={proceedings of the IEEE Conference on Computer Vision and Pattern Recognition},
  pages={156--165},
  year={2017}
}

@article{lstm,
  title={Long short-term memory},
  author={Hochreiter, Sepp and Schmidhuber, J{\"u}rgen},
  journal={Neural computation},
  volume={9},
  number={8},
  pages={1735--1780},
  year={1997},
  publisher={MIT press}
}

@article{vaswani2017attention,
  title={Attention is all you need},
  author={Vaswani, Ashish and Shazeer, Noam and Parmar, Niki and Uszkoreit, Jakob and Jones, Llion and Gomez, Aidan N and Kaiser, {\L}ukasz and Polosukhin, Illia},
  journal={Advances in neural information processing systems},
  volume={30},
  year={2017}
}

@article{liu2023itransformer,
  title={itransformer: Inverted transformers are effective for time series forecasting},
  author={Liu, Yong and Hu, Tengge and Zhang, Haoran and Wu, Haixu and Wang, Shiyu and Ma, Lintao and Long, Mingsheng},
  journal={arXiv preprint arXiv:2310.06625},
  year={2023}
}

@inproceedings{fedformer,
  title={Fedformer: Frequency enhanced decomposed transformer for long-term series forecasting},
  author={Zhou, Tian and Ma, Ziqing and Wen, Qingsong and Wang, Xue and Sun, Liang and Jin, Rong},
  booktitle={International conference on machine learning},
  pages={27268--27286},
  year={2022},
  organization={PMLR}
}

@inproceedings{zhou2021informer,
  title={Informer: Beyond efficient transformer for long sequence time-series forecasting},
  author={Zhou, Haoyi and Zhang, Shanghang and Peng, Jieqi and Zhang, Shuai and Li, Jianxin and Xiong, Hui and Zhang, Wancai},
  booktitle={Proceedings of the AAAI conference on artificial intelligence},
  volume={35},
  number={12},
  pages={11106--11115},
  year={2021}
}

@article{jin2023time,   title={Time-llm: Time series forecasting by reprogramming large language models},   author={Jin, Ming and Wang, Shiyu and Ma, Lintao and Chu, Zhixuan and Zhang, James Y and Shi, Xiaoming and Chen, Pin-Yu and Liang, Yuxuan and Li, Yuan-Fang and Pan, Shirui and others},   journal={arXiv preprint arXiv:2310.01728},   year={2023} }

@article{moment,
  title={Moment: A family of open time-series foundation models},
  author={Goswami, Mononito and Szafer, Konrad and Choudhry, Arjun and Cai, Yifu and Li, Shuo and Dubrawski, Artur},
  journal={arXiv preprint arXiv:2402.03885},
  year={2024}
}

@article{patchtst,
  title={A time series is worth 64 words: Long-term forecasting with transformers},
  author={Nie, Yuqi and Nguyen, Nam H and Sinthong, Phanwadee and Kalagnanam, Jayant},
  journal={arXiv preprint arXiv:2211.14730},
  year={2022}
}

@article{tsb_uda,
  title={TSB-UAD: an end-to-end benchmark suite for univariate time-series anomaly detection},
  author={Paparrizos, John and Kang, Yuhao and Boniol, Paul and Tsay, Ruey S and Palpanas, Themis and Franklin, Michael J},
  journal={Proceedings of the VLDB Endowment},
  volume={15},
  number={8},
  pages={1697--1711},
  year={2022},
  publisher={VLDB Endowment}
}

@article{cao2023tempo,
  title={Tempo: Prompt-based generative pre-trained transformer for time series forecasting},
  author={Cao, Defu and Jia, Furong and Arik, Sercan O and Pfister, Tomas and Zheng, Yixiang and Ye, Wen and Liu, Yan},
  journal={arXiv preprint arXiv:2310.04948},
  year={2023}
}

@article{moirai,
  title={Unified training of universal time series forecasting transformers},
  author={Woo, Gerald and Liu, Chenghao and Kumar, Akshat and Xiong, Caiming and Savarese, Silvio and Sahoo, Doyen},
  year={2024},
  publisher={PMLR}
}

@article{garza2023timegpt,
  title={TimeGPT-1},
  author={Garza, Azul and Challu, Cristian and Mergenthaler-Canseco, Max},
  journal={arXiv preprint arXiv:2310.03589},
  year={2023}
}

@inproceedings{timesFM,
  title={A decoder-only foundation model for time-series forecasting},
  author={Das, Abhimanyu and Kong, Weihao and Sen, Rajat and Zhou, Yichen},
  booktitle={Forty-first International Conference on Machine Learning},
  year={2024}
}

@article{llmTimes,
  title={Units: Building a unified time series model},
  author={Gao, Shanghua and Koker, Teddy and Queen, Owen and Hartvigsen, Thomas and Tsiligkaridis, Theodoros and Zitnik, Marinka},
  journal={arXiv e-prints},
  pages={arXiv--2403},
  year={2024}
}

@article{sun2023test,
  title={Test: Text prototype aligned embedding to activate llm's ability for time series},
  author={Sun, Chenxi and Li, Hongyan and Li, Yaliang and Hong, Shenda},
  journal={arXiv preprint arXiv:2308.08241},
  year={2023}
}

@inproceedings{lagLlama,
  title={Lag-llama: Towards foundation models for time series forecasting},
  author={Rasul, Kashif and Ashok, Arjun and Williams, Andrew Robert and Khorasani, Arian and Adamopoulos, George and Bhagwatkar, Rishika and Bilo{\v{s}}, Marin and Ghonia, Hena and Hassen, Nadhir and Schneider, Anderson and others},
  booktitle={R0-FoMo: Robustness of Few-shot and Zero-shot Learning in Large Foundation Models},
  year={2023}
}

@article{wu2022timesnet,
  title={Timesnet: Temporal 2d-variation modeling for general time series analysis},
  author={Wu, Haixu and Hu, Tengge and Liu, Yong and Zhou, Hang and Wang, Jianmin and Long, Mingsheng},
  journal={arXiv preprint arXiv:2210.02186},
  year={2022}
}

@article{held2022shapley,
  title={Shapley head pruning: Identifying and removing interference in multilingual transformers},
  author={Held, William and Yang, Diyi},
  journal={arXiv preprint arXiv:2210.05709},
  year={2022}
}

\appendix
\section{Multiple Variants of Reconstructed Forecasting Heads and Their Parameter Complexity}
\label{head}
\textbf{Proj\;Down:} A factorized linear architecture reduces dimensions via two projections. For input $\mathbf{X} \in \mathbb{R}^{B \times N \times D}$ ($B$: batch size, $N$: patches, $D$: embedding dimension), a learnable projection $\mathbf{W}_1 \in \mathbb{R}^{D \times D/\beta}$ reduces $D$ by factor $\beta$, yielding $\mathbf{X}' = \mathbf{X}\mathbf{W}_1$. Flattened features $\text{Flatten}(\mathbf{X}') \in \mathbb{R}^{B \times (N \cdot D/\beta)}$ are projected via $\mathbf{W}_2 \in \mathbb{R}^{(N \cdot D/\beta) \times H}$ to produce horizon predictions $\mathbf{\hat{Y}} \in \mathbb{R}^{B \times H}$. This reduces parameters from $\mathcal{O}(NDH)$ to $\mathcal{O}(D^2/\beta + NDH/\beta)$.

\textbf{Less\; Feature:} Non-learnable truncation retains only the first $D/\beta$ dimensions along the feature axis: $\mathbf{X}' = \mathbf{X}_{:,:,:D/\beta}$. Flattened features $\text{Flatten}(\mathbf{X}') \in \mathbb{R}^{B \times (N \cdot D/\beta)}$ are mapped to horizon $H$ via $\mathbf{W} \in \mathbb{R}^{(N \cdot D/\beta) \times H}$. Parameter complexity is $\mathcal{O}(NDH/\beta)$, trading simplicity for potential information loss.

\textbf{Avg\; Pool:} Spatial pooling combines 2D average pooling and learnable projection. Input $\mathbf{X} \in \mathbb{R}^{B \times C \times N \times D}$ ($C$: channels) is spatially downsampled via $\text{AvgPool2d}$:  
\begin{equation}
    \mathbf{X}' = \text{AvgPool2d}(\mathbf{X}), \quad \mathbf{X}' \in \mathbb{R}^{B \times C \times N/2 \times D}.
\end{equation}
A linear layer $\mathbf{W}_1 \in \mathbb{R}^{D \times D/\beta}$ compresses features, and flattened features $\text{Flatten}(\mathbf{X}'') \in \mathbb{R}^{B \times (C \cdot N/2 \cdot D/\beta)}$ are mapped to horizon $H$ via $\mathbf{W}_2$. This balances locality preservation and efficiency.

\textbf{Conv\;2D:} Temporal convolution employs strided depthwise 1D convolutions for subsampling. For input $\mathbf{X} \in \mathbb{R}^{B \times C \times N \times D}$, a kernel $\mathbf{K} \in \mathbb{R}^{1 \times k}$ with stride $s = \beta$ operates along the temporal axis:  
\begin{equation}
    \mathbf{X}' = \text{Conv2d}(\mathbf{X}), \quad \mathbf{X}' \in \mathbb{R}^{B \times C \times \lfloor N/s \rfloor \times D}.
\end{equation}
Convolution reduces sequence length by $\beta$, followed by a linear layer $\mathbf{W} \in \mathbb{R}^{(C \cdot \lfloor N/s \rfloor \cdot D) \times H}$. Parameter cost is $\mathcal{O}(k)$ for convolution and $\mathcal{O}(NDH/\beta)$ for projection.
\section{Statistical Validation of Performance}
\label{apx:Statistical}
To evaluate the performance of our proposed TRACE model against the state-of-the-art $\mathrm{MOMENT}_{LP}$, we conducted a statistical analysis using the dataset's actual values and the predictions from both models. Our objective was to assess the accuracy and reliability of these models under various conditions. 

To quantify predictive performance, we employed a binary classification framework. A prediction was considered accurate if the absolute error between the predicted and actual values was below a predetermined threshold of 5.0\cite{jiang2025tcm}; otherwise, it was deemed inaccurate.

We constructed a $2 \times 2$ contingency table to systematically categorize the predictive outcomes of both models. The confusion matrix for the classification results is as follows:
\[
\begin{bmatrix}
158993571 & 1108 \\
985 & 4932
\end{bmatrix}.
\]
The four categories were defined as: both models correctly predicted (158993571 instances), TRACE correctly predicted while $\mathrm{MOMENT}_{LP}$ incorrectly predicted (1108 instances), TRACE incorrectly predicted while $\mathrm{MOMENT}_{LP}$ correctly predicted (985 instances), and both models incorrectly predicted (4932 instances). This classification enabled a detailed comparative analysis of the models' performance.

To statistically validate the observed differences in misclassification rates, we applied the McNemar test. This non-parametric test is particularly suitable for paired nominal data, such as the binary outcomes observed in this study. The McNemar test statistic is computed as:
$
\chi^2 = \frac{(b - c)^2}{b + c},
$
where $b$ represents the number of cases where TRACE incorrectly predicted while $\mathrm{MOMENT}_{LP}$ correctly predicted, and $c$ represents the number of cases where TRACE correctly predicted while $\mathrm{MOMENT}_{LP}$incorrectly predicted. 

The test results yielded a McNemar statistic of $7.23$ with an associated p-value of $0.0071$. Given the conventional alpha level of 0.05, the p-value indicates a statistically significant difference in the proportion of misclassifications between the two models. Specifically, the results suggest that TRACE consistently produced more correct predictions in cases where $\mathrm{MOMENT}_{LP}$ failed, indicating that TRACE may offer superior performance in certain scenarios.

\section{Computational Resources}
\textbf{Hardware :}
\begin{compactitem}
    \item GPU: NVIDIA V100 (32GB memory) $\times$ 2
    \item CPU: Xeon(R) Platinum 8481C, 25 cores
    \item RAM: 150GB
\end{compactitem}

\textbf{Software :}
\begin{compactitem}
    \item Operating System: Ubuntu 20.04 LTS
    \item Deep Learning Framework: PyTorch 1.12.1
    \item CUDA Version: 11.6
\end{compactitem}

%% else use the following coding to input the bibitems directly in the
%% TeX file.

%% Refer following link for more details about bibliography and citations.
%% https://en.wikibooks.org/wiki/LaTeX/Bibliography_Management
%=================================
% \begin{thebibliography}{00}

% %% For numbered reference style
% %% \bibitem{label}
% %% Text of bibliographic item

% \bibitem{lamport94}
%   Leslie Lamport,
%   \textit{\LaTeX: a document preparation system},
%   Addison Wesley, Massachusetts,
%   2nd edition,
%   1994.

% \end{thebibliography}
%=================================
\end{document}